\documentclass{article}

\PassOptionsToPackage{table}{xcolor}

\usepackage{microtype}
\usepackage{graphicx}
\usepackage{subcaption}
\usepackage{booktabs} %

\usepackage{hyperref}

\usepackage[accepted]{icml2026}

\usepackage{amsmath,amsfonts,bm}
\usepackage{bbm}
\usepackage[mathscr]{eucal}

\def\1{\bm{1}}

\def\<{\langle}
\def\>{\rangle}

\def\mA{{\bm{A}}}

\def\mD{{\bm{D}}}
\def\mE{{\bm{E}}}

\def\mH{{\bm{H}}}
\def\mI{{\bm{I}}}

\def\mL{{\bm{L}}}

\def\mS{{\bm{S}}}

\def\mV{{\bm{V}}}

\DeclareMathAlphabet{\mathsfit}{\encodingdefault}{\sfdefault}{m}{sl}
\SetMathAlphabet{\mathsfit}{bold}{\encodingdefault}{\sfdefault}{bx}{n}

\let\hat\relax
\newcommand{\hat}{\widehat}

\usepackage{nicefrac}
\usepackage{multirow}
\usepackage{mathabx}
\usepackage{float}
\usepackage{colortbl}
\usepackage{enumitem}
\usepackage{adjustbox}
\usepackage{afterpage}
\usepackage{scalerel,xparse}
\usepackage{capt-of}
\usepackage{etoolbox}
\usepackage{wrapfig}

\usepackage[table]{xcolor}

\definecolor{imp1}{RGB}{230,245,255} %
\definecolor{imp2}{RGB}{198,219,239}
\definecolor{imp3}{RGB}{158,202,225}
\definecolor{imp4}{RGB}{107,174,214} %

\icmltitlerunning{Relighting as a Probe of Visual Priors via Augmented Latent Intrinsics}

\begin{document}

\twocolumn[
  \icmltitle{Relighting as a Probe of Visual Priors via Augmented Latent Intrinsics}

  \icmlsetsymbol{equal}{*}

  \begin{icmlauthorlist}
    \icmlauthor{Xiaoyan Xing}{aff1}
    \icmlauthor{Xiao Zhang}{aff2}
    \icmlauthor{Sezer Karagolu}{aff1}
    \icmlauthor{Theo Gevers}{aff1}
    \icmlauthor{Ananad Bhattad}{aff3}
  \end{icmlauthorlist}

  \icmlaffiliation{aff1}{UvA-Bosch Delta Lab, University of Amsterdam, Amsterdam, Netherlands}
  \icmlaffiliation{aff2}{The University of Chicago, Chicago, USA}
  \icmlaffiliation{aff3}{Johns Hopkins University, Baltimore, USA}

  \icmlcorrespondingauthor{Xiaoyan Xing}{x.xing@uva.nl}

  \icmlkeywords{Machine Learning, Image Relighting, Visual Representations, ICML}

  \vskip 0.3in
  {\begin{center}
\centering
\captionsetup{type=figure}
\vspace{-1em}%
\includegraphics[width=\linewidth]{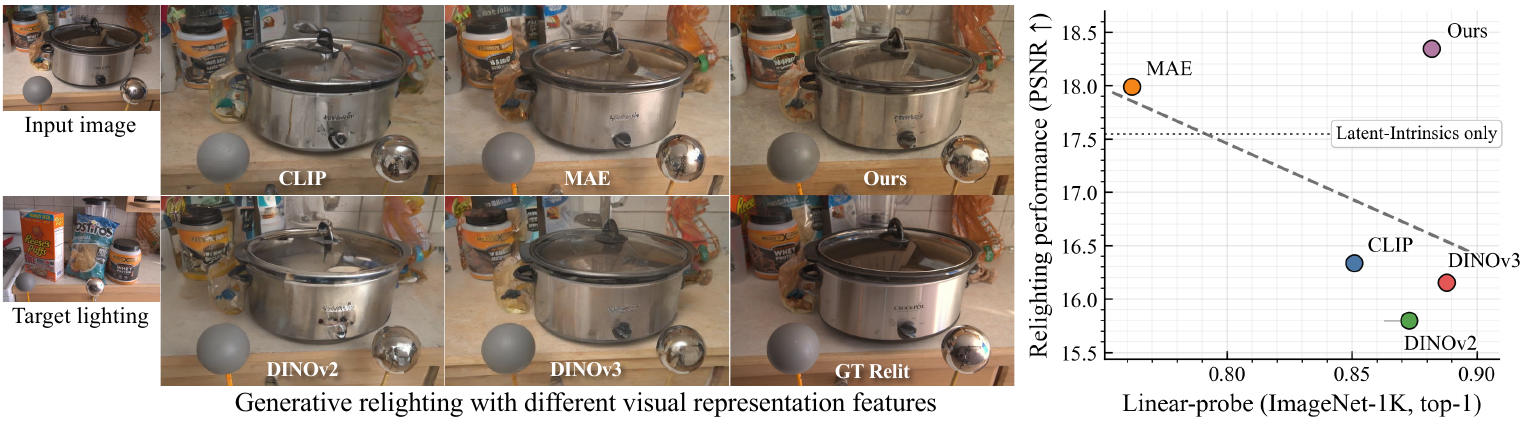}
  \vspace{-5pt}
  \caption{\textbf{Stronger Semantic Encoders Can Harm Relighting Performance.}
  \textbf{Left:} Visual comparison on a scene with complex specular materials. The task is to relight the input image (top-left) using the target illumination (bottom-left), which requires moving specular highlights from left to right, as indicated by the chrome sphere. While features from semantic encoders (CLIP, DINO) fail to reproduce realistic highlights, the MAE plausibly moves the highlight but blurs fine details, such as text labels. Our method (top-right), which combines features from RADIO (a pretrained model; distilled from many vision encoders)  with latent intrinsics, closely matches the ground truth.
  \textbf{Right:} Quantitative analysis reveals a trade-off: for most encoders optimized for pure semantics, relighting quality (PSNR) is inversely correlated with recognition performance ({ImageNet-1K linear probing as reported in the original papers.}). Our approach breaks this trend, achieving high performance on both tasks.}
  \label{fig:teaser}
\end{center}%
\vspace{1em}
}
]

\printAffiliationsAndNotice{} %

\begin{abstract}
Image-to-image relighting requires representations that separate illumination from scene properties while preserving dense geometry, material, and photometric cues. We use this task as a probe of visual priors: unlike recognition tasks that reward invariance, relighting tests whether visual features retain the information needed for light transfer. Through a controlled generative relighting framework, we find that strong semantic encoders can degrade relighting quality, exposing a semantic--photometric trade-off between abstraction and physical fidelity. We introduce \textbf{Augmented Latent Intrinsics (ALI)}, which balances this trade-off by fusing dense, pixel-aligned visual features into a latent-intrinsic relighting model and refining it with self-supervision on unlabeled real image pairs. ALI improves relighting quality, especially on glossy, metallic, and transparent materials, and demonstrates that generative relighting is an effective tool for quantifying what visual encoders encode about the physical world.
\vspace{-10pt}
\end{abstract}

\section{Introduction}

Image-based relighting is the task of transferring illumination from one image to another. It has important applications in augmented reality, computational photography, and digital content creation. Yet it remains fundamentally ill-posed: brightness may arise from a light source or a shiny surface, and darkness may indicate either shadow or material absorption. Successful relighting, therefore, requires more than semantic recognition. It demands visual representations that encode how illumination interacts with scene geometry, material properties, and fine-grained appearance.

In this paper, we use generative relighting as a probe of visual representations. While recognition tasks often reward invariance to lighting, texture, and local appearance, relighting requires exactly the information that such invariances tend to suppress. A representation useful for relighting must preserve dense, pixel-aligned photometric structure while still providing enough semantic context to respect object and material boundaries. This makes relighting a controlled testbed for studying what visual encoders encode about the physical world: not merely whether they recognize objects, but whether their features support physically meaningful image transformations.

We focus on image-to-image relighting as a practical setting for this analysis. Unlike classical inverse-graphics pipelines that explicitly estimate geometry, reflectance, and light sources, we test whether learned visual features---augmented with latent intrinsic representations~\citep{Latent-Intrinsic}---are sufficient to support illumination transfer. This direct formulation avoids explicit decomposition into physical components, but places strong demands on the representation. Recent diffusion-based relighting models show impressive generative capacity, yet often struggle with materials involving complex, view-dependent effects such as specularity and transparency~\citep{LumiNet, zeng2024rgb, DiffusionRenderer2025, he2025unirelight}. These failures suggest that generative power alone is insufficient: the conditioning representation must also encode the physical cues needed for light transport.

A natural hypothesis is that large-scale pretrained visual encoders, such as CLIP and DINO, should provide strong priors for resolving these ambiguities. Our findings reveal a more nuanced picture. Features optimized for semantic invariance through contrastive learning or self-distillation can systematically degrade relighting performance, despite their strong recognition ability. In contrast, features from a masked auto-encoder (MAE), trained with a dense pixel reconstruction objective, yield better relighting results even though they are weaker semantic representations (Fig.~1). This exposes a critical mismatch between semantic representation quality and physical usefulness: the inductive biases that make features robust for recognition can discard the fine-grained photometric and spatial information required for relighting.

This motivates our central question of investigation in this paper: \emph{what makes a visual representation relightable?} We define relightability operationally and objectively as the ability to support three properties: (i) photometric consistency under light transfer, (ii) material fidelity for view- and material-dependent effects, and (iii) illumination robustness while retaining dense per-pixel detail. Through controlled comparisons of different pretrained encoders within the same generative relighting pipeline, we show that relightability is controlled by a semantic--photometric trade-off. High-level semantic features provide object-level context but often lose dense appearance cues; low-level features preserve photometric detail but lack semantic structure. Dense, pixel-aligned encoders such as RADIO~\citep{heinrich2024radio}, which distill complementary visual teachers, best balance these requirements.

To build this probe, we introduce \textbf{Augmented Latent Intrinsics (ALI)}, a controlled framework for image-to-image relighting that allows us to quantify how different visual priors affect generative light transfer. ALI combines latent intrinsic features with semantic features from a frozen visual encoder through a lightweight fusion adapter, and aligns the resulting representation with a diffusion-based decoder using a progressive training schedule. This design enables systematic analysis of how encoder choice changes relighting behavior while also producing a practical relighting model. A final self-supervision stage improves robustness on in-the-wild images without altering the core representation analysis.

Experiments on the MIIW benchmark show that ALI achieves state-of-the-art performance among open-sourced diffusion-based relighting methods, with especially large gains on glossy and specular materials. More importantly, our analysis demonstrates that relighting can serve as a sensitive probe of physically grounded visual priors: encoders that appear strong under semantic benchmarks may fail to preserve the material, geometric, and photometric cues required for realistic illumination transfer.

In summary, our contributions are:
\vspace{-10pt}
\begin{itemize}[leftmargin=1em,itemsep=-5pt]
\item We propose generative relighting as a probe of physically grounded visual priors, using relightability to quantify what visual encoders encode about illumination, material, and geometry.
\item We introduce \textbf{ALI}, a controlled image-to-image relighting framework that exposes and balances the semantic--photometric trade-off in pretrained visual representations.
\item We provide a systematic analysis showing that strong semantic encoders can fail at relighting, while dense pixel-aligned representations better preserve the physical cues required for light transfer.
\item ALI achieves state-of-the-art results on MIIW among open-sourced diffusion-based methods, improving RMSE by \textbf{4.5\%} and SSIM by \textbf{4.9\%} over \textit{LumiNet}, with the largest gains on glossy and specular materials.
\end{itemize}

\section{Related Work}
\noindent\textbf{Generative Relighting.}
Image-based relighting modifies illumination while preserving scene content, implicitly requiring a model to separate lighting from geometry, material, and semantics. Classical inverse-rendering methods address this ambiguity by estimating or assuming physical factors such as geometry, reflectance, and illumination~\citep{indoor_3d_4,indoor_3d_1,indoor_3d_2,indoor_3d_3,indoor_3d_5,garon2019fast,gardner2019deep}, but often rely on controlled assumptions and accurate intermediate estimates.

Recent generative methods learn relighting directly, achieving compelling results for objects~\citep{zeng2024dilightnet,FlashTex,Neural-Garffer,IC-Light,bharadwaj2024genlit}, portraits~\citep{Pandey:2021,he2024diffrelight,mei2025lux}, and indoor scenes~\citep{kocsis2024lightit,xing2024retinex,choi2024scribblelight,zeng2024rgb}. However, strong systems such as the concurrent \textit{LightLab}~\citep{LightLab}, \textit{UniRelight}~\citep{he2025unirelight}, and \textit{IntrinsicEdit}~\citep{IntrinsicEdit} rely heavily on dense supervision, synthetic data, or explicit physical annotations. Unsupervised alternatives exploit priors from pretrained GANs and diffusion models~\citep{StyLitGAN,LumiNet}, but photometric objectives alone remain vulnerable to intrinsic ambiguities, especially for glossy, transparent, or spatially complex materials. We instead treat relighting as a controlled probe for evaluating whether learned visual priors preserve the physical information needed for light transfer.

\noindent\textbf{Physically Grounded Visual Priors.}
Intrinsic image decomposition~\citep{barrowtenenbaum} provides a natural physical prior by factorizing images into illumination-independent properties, such as albedo, and illumination-dependent components, such as shading. Prior work has used low-level cues~\citep{baslamisli2021shadingnet,luo2020niid,das2022pie,fan2018revisiting,chen2013simple,xing2022point}, physical assumptions~\citep{barron2014shape,grosse2009ground}, ordinal supervision~\citep{careaga2023intrinsic,careaga2024colorful,dilleIntrinsicHDR}, semantic reasoning~\citep{baslamisli2018joint}, and generative priors~\citep{bhattad2024stylegan,du2023generative,kocsis2024iid,zeng2024rgb,luo2024intrinsicdiffusion,chen2024intrinsicanything}. Self-supervised methods further exploit real image pairs captured under varying illumination~\citep{li2018learning,ma2018single,janner2017self}.

These representations preserve dense photometric structure, but intrinsic decomposition alone does not resolve all relighting ambiguities. In real scenes, distinguishing lighting from material or object identity often requires semantic context. \textit{Latent Intrinsics}~\citep{Latent-Intrinsic} captures useful photometric structure in latent space without explicit labels, but lacks the semantic grounding needed for complex materials and object boundaries. We therefore use latent intrinsics as one component in a broader probe of physically grounded visual priors.

\noindent\textbf{Semantic Visual Representations.}
Large-scale visual encoders trained with contrastive, distillation, or reconstruction objectives~\citep{he2020momentum,chen2020simple,caron2021emerging,vincent2008extracting,ho2020denoising,he2022masked,zhang2024residual,zhang2024deciphering} encode rich information about object identity, layout, parts, and material categories. Such semantics are useful for relighting, where object boundaries and material identity help constrain illumination transfer.

However, semantic strength does not guarantee physical usefulness. Many high-level encoders are optimized for invariance to lighting, texture, and local appearance, while relighting requires these signals to be preserved. This creates a semantic--photometric trade-off: recognition-oriented features may lose dense light-transfer cues, while photometric features may lack semantic structure. Relighting therefore offers a sensitive probe of visual representations, requiring features that encode both \emph{what} the scene contains and \emph{how} its surfaces respond to light.

To our knowledge, prior work has not systematically examined large-scale visual encoders through generative relighting. We address this with \textbf{ALI}, a controlled image-to-image relighting framework for quantifying how visual priors affect relighting behavior. Our analysis shows that dense, pixel-aligned representations such as RADIOv2.5H~\citep{heinrich2024radio}, when fused with latent intrinsic features, better balance semantic grounding and photometric fidelity than either prior alone.

\begin{figure*}[t!]
    \centering
    \includegraphics[width=0.7\linewidth]{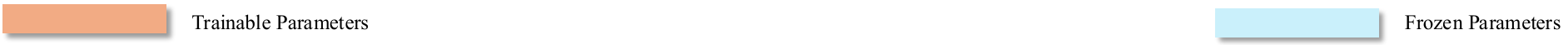}
    \includegraphics[width=0.98\linewidth]{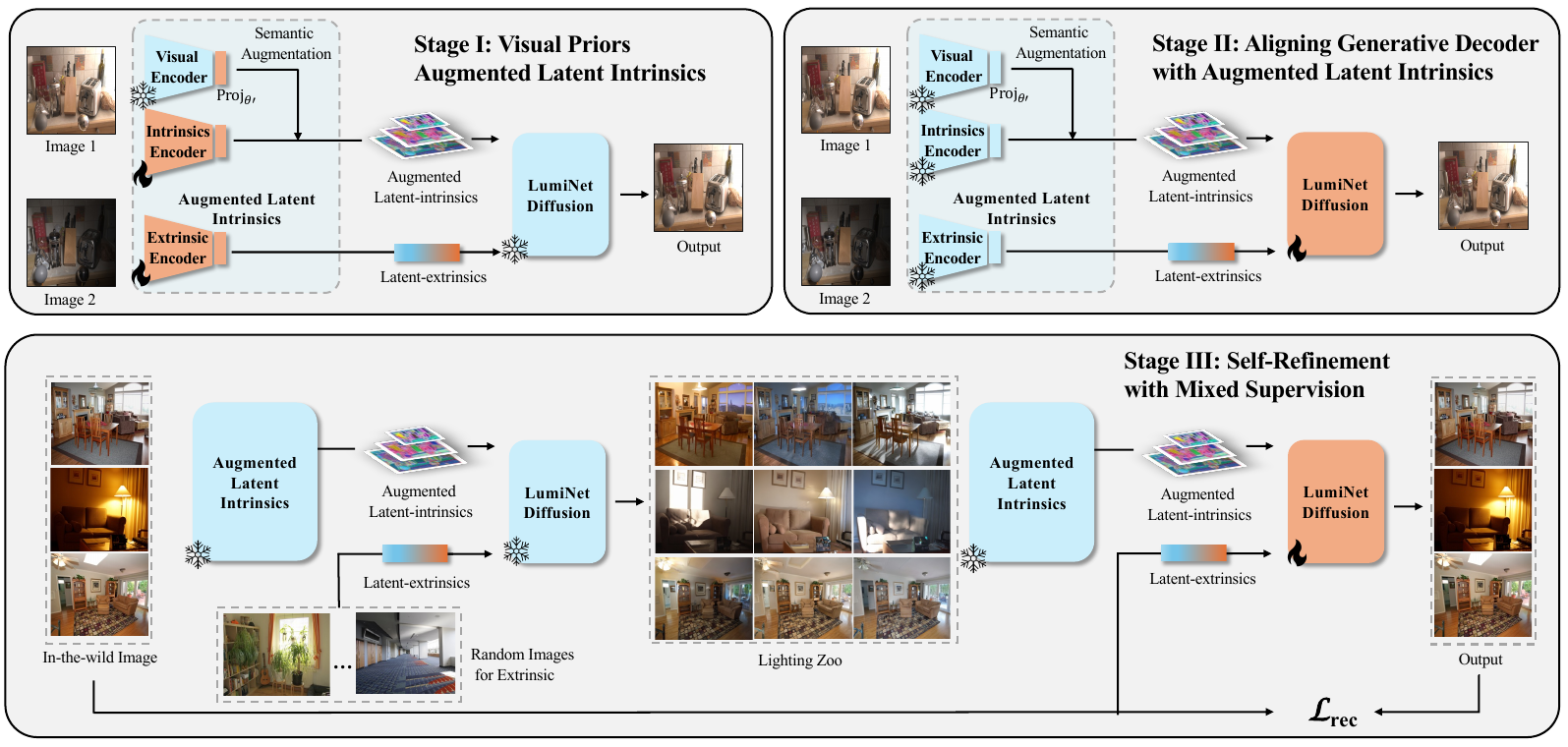}
    \caption{\textbf{Our Three-Stage Training Pipeline for Augmented Latent Intrinsics (ALI).} Our method progressively adapts a pretrained visual encoder and fine-tunes a generative decoder for high-fidelity, unsupervised relighting. \textbf{Stage I: Augmenting Latent Intrinsics.} We inject semantic features from a frozen vision encoder into the intrinsics encoder. This creates our semantically-enriched \textbf{ALI}, which better disentangles scene properties from illumination. \textbf{Stage II: Aligning the Generative Decoder.} With the encoder fixed, we fine-tune the LumiNet diffusion decoder to condition on the new ALI representation, aligning the generator with the scene's improved physical understanding. \textbf{Stage III: Self-Refinement.} We generate a ``Lighting Zoo'' ({examples in supplementary}) of pseudo-relit images to overcome data scarcity. These synthetic images serve as new inputs, with the original real image as the ground truth. This self-supervision trains the network to ignore artifacts and focus on essential structural properties, improving realism for in-the-wild images.}
    \label{fig:pipeline}
    \vspace{-5pt}
\end{figure*}

 \section{Preliminaries}
\label{sec:preliminaries}

Our work builds on the concept of latent intrinsics, where lighting-invariant features can be learned from multi-illumination image pairs without direct supervision~\citep{Latent-Intrinsic}. Given an image $\mI_s^l$ of a scene $s$ under lighting $l$, an encoder $\mE_{\theta}$ disentangles it into a set of hierarchical, lighting-invariant intrinsic features $\{\mS_{s,i}^{l}\}$ and a global lighting embedding $\mL_s^{l}$. A decoder $\mD_{\phi}$ can then relight the scene by combining the intrinsics from one view with the lighting from another:
\begin{equation}
    \Tilde{\mI}_s^{l_1\rightarrow l_2} = \mD_{\phi}(\{\mS_{s,i}^{l_1}\}, \mL_s^{l_2}).
\end{equation}
The model is trained from scratch using a combination of reconstruction, intrinsic invariance, and latent space regularization losses. However, because these models are typically trained on a limited set of real-world multi-illumination pairs, their learned representations are constrained to the lighting and material types seen during training. This often leads to failures in accurately disentangling complex materials under diverse, unseen lighting conditions. To address this, we propose to augment the latent intrinsic features with powerful visual priors from foundation models trained on large-scale, diverse image collections. While these priors are effective for high-level tasks, their utility for the fine-grained physical reasoning required in relighting remains unexplored.

\section{Method}
\label{sec:method}

Our goal is to use image-to-image relighting as a controlled probe of physically grounded visual priors. Rather than changing the full relighting pipeline, we keep the backbone, training data, and decoder objective fixed, and vary the pretrained visual encoder used to augment the latent intrinsic representation. This isolates how different visual priors affect relightability, including photometric consistency, material fidelity, and preservation of dense scene structure.

We instantiate this probe with \textbf{Augmented Latent Intrinsics (ALI)}, a framework built on LumiNet~\citep{LumiNet}. ALI injects frozen visual encoder features into latent intrinsic representations and aligns them with a generative relighting decoder through a three-stage training strategy.

\noindent\textbf{Stage 1: Augmenting Latent Intrinsics.} To improve the disentanglement of latent intrinsic (like albedo) and extrinsic (like lighting), we inject semantic context from a frozen visual encoder $\mE_{\text{sem}}$ (e.g., RADIOv2.5) into the latent intrinsics. First, we extract a hierarchy of feature maps $\{\mV_{s,i}^l\}_{i=1}^N = \mE_{\text{sem}}(\mI_s^l)$ from an input image $\mI_s^l$, {where $N$ denotes the number of selected intermediate layers from each VRM encoder; the exact layer indices and feature dimensions are provided in the supplementary material}. These maps are upsampled to the input resolution and concatenated into a pixel-wise hypercolumn descriptor $\mH_s^l(x, y)$. A learnable projection layer $\text{Proj}_{\theta'}$ then aligns these features with the intrinsic features $\mS_{s,i}^l$ from the relighting encoder $\mE_\theta$:
\begin{equation}
    \mA_{s,i}^l = \text{Proj}_{\theta'}(\mH_{s,i}^l) + \mS_{s,i}^l.
\end{equation}
In this stage, we freeze the visual encoder $\mE_{\text{sem}}$ and all decoders, training only the relighting encoder $\mE_\theta$ and the projection layers $\text{Proj}_{\theta'}$. The training objective combines a standard reconstruction loss $\mathcal{L}_\text{relight}$ and a hyperspherical regularization loss $\mathcal{L}_\text{reg}$ applied to both intrinsic and lighting features~\citep{Latent-Intrinsic}:
{\begin{align}
    \mathcal{L}_\text{reg}(\mA) &= \left\| R(\mA) - R(\hat{\mA}) \right\|_2^2 \\
    R(\mA) &= \log \det \left( \mI + \frac{d}{n\lambda^2} \mA^\top \mA \right),
\end{align}
where $\hat{\mA}$ is sampled from a uniform hyperspherical distribution, $n$ is the number of spatial locations, $d$ is the feature dimension, and $\lambda$ is a regularization temperature parameter. This term encourages features to uniformly spread out in the sphere, aiding feature optimization.}

To stabilize training, we replace the original intrinsic invariance loss with an improved version that regularizes the intrinsic features $\mA^{l_n}_{s,i}$ of a scene $s$ toward their mean across $M$ different lighting conditions:
\begin{equation}
    \mathcal{L}_{\text{Improved Intrinsics}} = \sum_{s,m}\|\mA^{l_m}_{s,i} - \frac{1}{M}\sum_{m'}\mA^{l_{m'}}_{s,i}\|_2.
\end{equation}
The total loss for this stage is $\mathcal{L}_\text{\textrm{Stage1}} = \mathcal{L}_{\text{relight}}+\mathcal{L}_{\text{Improved Intrinsics}} + \sum_i \mathcal{L}_{\text{reg}}(\mA_{s,i}^l) + \mathcal{L}_{\text{reg}}(\mL^l_s)$.

\noindent\textbf{Stage 2: Aligning the Generative Decoder.} {With visual encoder $\mE_\text{sem}$ relighting encoder $E_\theta$ and projection layer $\text{Proj}_{\theta'}$ fixed}, we fine-tune LumiNet's diffusion decoder $\mD_{\phi'}$ to align with our semantically-augmented latent intrinsic representation $\mA_{s,i}^{l_1}$. The decoder is trained to predict the noise $\epsilon$ added to the target lighting latent $\mL_s^{l_2}$ at timestep $t$, conditioned on the new intrinsics. The optimization uses only the standard denoising score-matching loss from DDPM~\citep{ho2020denoising}:
\begin{equation}
\vspace{-2pt}
\mathcal{L}_{\text{Stage2}}=\mathbb{E}_{t,\epsilon}\left[\|\mD_{\phi'}(\alpha_t\mL_s^{l_2}+\beta_t\epsilon,\{\mA_{s,i}^{l_1}\},\mL_s^{l_2})-\epsilon\|_2^2\right]
\end{equation}

\noindent\textbf{Stage 3: Self-Refinement.}
To address the scarcity of paired real-world data, we refine the decoder using a self-training scheme. Pseudo-relit pairs are generated by transferring illumination between randomly sampled images within a batch. The decoder is fine-tuned on these pseudo-pairs using the same diffusion loss as in Stage~II, while periodically mixing in same-image reconstructions (source equals target) to preserve content fidelity. This self-refinement improves realism and robustness without requiring any labeled data.

\section{Experiments}
\noindent\textbf{Experimental Setup.}
We train \textbf{ALI} on real image pairs from two datasets: the \textit{MIT Multi-Illuminant} (MIIW) dataset~\citep{MIIW}, which contains 985 scenes captured under 25 lighting conditions, and the \textit{BigTime} dataset~\citep{li2018learning}, which contains 460 scenes under 20--50 natural illumination conditions.

All stages are trained with AdamW~\citep{loshchilov2017decoupled_adamw} using a learning rate of $4\times10^{-5}$. In \textbf{Stage I}, we use a scene-aware batch sampler that groups multiple illuminations of the same scene to enforce intrinsic consistency. In \textbf{Stage II}, this constraint is relaxed and images are sampled across scenes to increase decoder diversity. In \textbf{Stage III}, we perform self-refinement using pseudo-relit data generated from a lighting zoo, randomly sampling 1,000 scenes for this process (Sec.~\ref{sec:lightzoo}).

At inference time, although ALI is trained on same-scene image pairs, it generalizes to arbitrary unpaired image-to-image relighting. Intrinsic features are extracted from a content image, while lighting embeddings are extracted from a separate illumination image and transferred through the diffusion decoder. Following LumiNet~\citep{LumiNet}, we use the bypass decoder~\citep{quadprior} by default to preserve identity. However, all feature-fusion experiments disable this bypass to isolate the effect of the augmented latent intrinsic representation.

\noindent\textbf{Evaluation Overview.}
We evaluate ALI in two complementary roles: as an image-to-image relighting model and as a controlled probe of physically grounded visual priors. First, we compare relighting quality against existing methods on MIIW and in-the-wild images. We then use the same framework to analyze how different pretrained visual encoders affect relightability, with emphasis on material-dependent behavior and the semantic--photometric trade-off.

Standard full-reference metrics such as PSNR, SSIM, and RMSE correlate imperfectly with perceived lighting quality and often emphasize low-frequency tone consistency over directional shadows or specular effects~\citep{giroux2024towards,LumiNet}. We therefore report them for protocol consistency and controlled ablations, while also relying on qualitative comparisons and human evaluation to assess physical plausibility.

\begin{figure*}
    \centering
    \includegraphics[width=\linewidth]{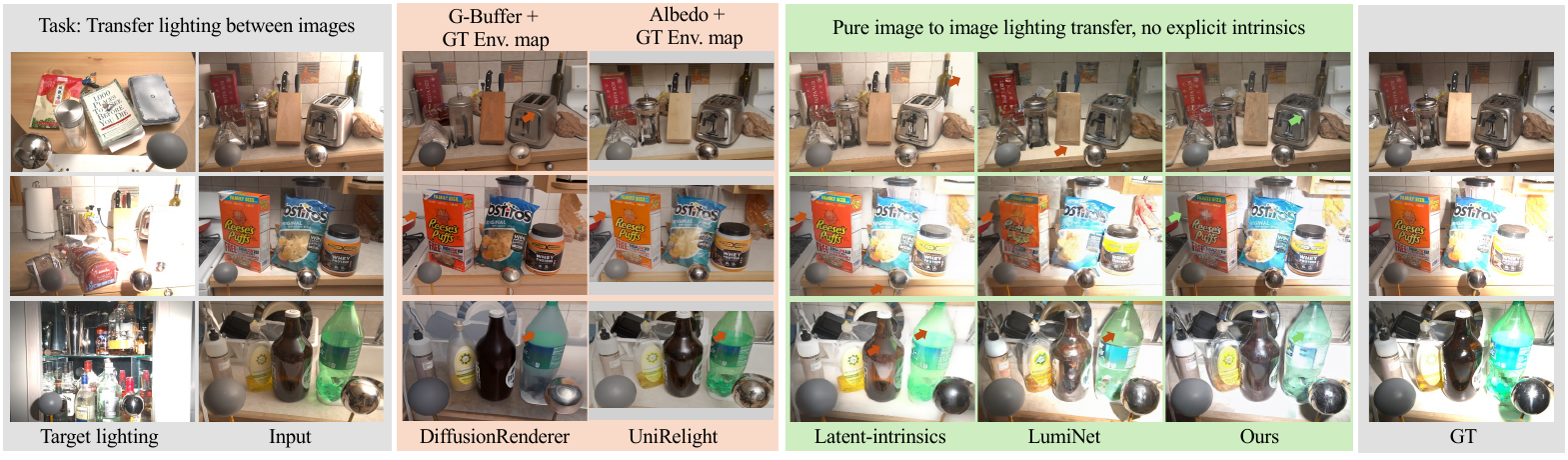}
    \vspace{-12pt}
    \caption{
    \textbf{Qualitative comparison of relighting methods on challenging MIIW test scenes.}
    The task is to relight the \emph{Input} scene using illumination from the \emph{Target lighting} image.
    Compared with competing methods, including approaches that rely on privileged information such as ground-truth light maps, G-buffers, albedo, or environment maps, our method produces more physically plausible results.
    In the \textbf{top row}, baselines render the metallic toaster with faint or blurry highlights, while our method produces sharper reflections closer to the ground truth.
    In the \textbf{second row}, baselines miss or distort the shadow of the orange cereal box, and \texttt{LumiNet} also blurs text on the packaging.
    In the \textbf{third row}, baselines struggle with transparency and caustic-like effects on the bottles, while our method produces a more plausible result.
    Red and green markers highlight representative failures and successes, respectively.
    Results for \texttt{UniRelight} were provided by the authors.
    \emph{Best viewed on screen with zoom.}
    }
    \label{fig:miiw_comparison}
\end{figure*}
\subsection{Relighting Performance}
\noindent\textbf{Relighting Performance on MIIW.}
We first evaluate cross-scene relighting on the unseen test split of MIIW. We compare against feed-forward methods, including SA-AE~\citep{hu2020sa} and Latent-Intrinsic~\citep{Latent-Intrinsic}, as well as diffusion-based methods, including RGB$\leftrightarrow$X~\citep{zeng2024rgb} and LumiNet~\citep{LumiNet}. Following the standard protocol~\citep{Latent-Intrinsic}, each trial samples one target image and 12 reference lighting conditions. We report results on both raw outputs and outputs after global color correction to compensate for white-balance discrepancies.

As shown in Tab.~\ref{tab:MIIW-cross}, ALI achieves state-of-the-art performance among open-sourced diffusion-based methods under both settings, improving SSIM by \textbf{4.9\%} and RMSE by \textbf{4.5\%} over LumiNet. Methods trained exclusively on MIIW, such as SA-AE and Latent-Intrinsic, can achieve stronger pixel-level scores, reflecting the bias of standard metrics toward tone consistency rather than accurate modeling of shadows, specularities, and material-dependent effects. This discrepancy is visible in qualitative comparisons (Fig.~\ref{fig:miiw_comparison}) and human evaluation (Tab.~\ref{tab:usr_study}); details of the human study are provided in the appendix.

Qualitatively, LumiNet produces coherent global illumination but often blurs material and geometric details, especially on reflective or transparent surfaces. In contrast, ALI better preserves fine structures and material-specific effects such as metallic highlights, cast shadows, and glass transparency. Several competing methods also use privileged inputs: DiffusionRenderer uses ground-truth light probes, while UniRelight requires albedo and environment maps. ALI achieves competitive or better relighting quality without 3D supervision, inverse-graphics labels, or ground-truth lighting maps.

\begin{figure*}[t]
\centering
\renewcommand{\arraystretch}{0.4}
\setlength{\tabcolsep}{1pt}
\resizebox{0.95\textwidth}{!}{
\begin{tabular}{ccccccc}

\rotatebox{90}{\small Light 1} &
\includegraphics[width=0.22\linewidth,height=15ex]{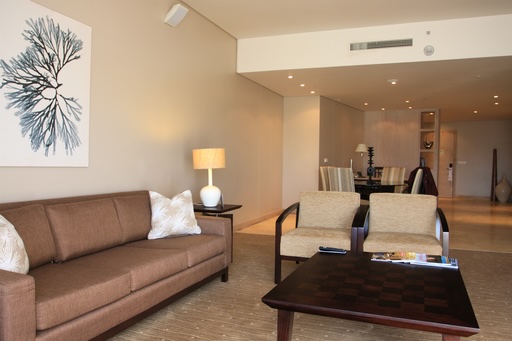} &
\includegraphics[width=0.22\linewidth,height=15ex]{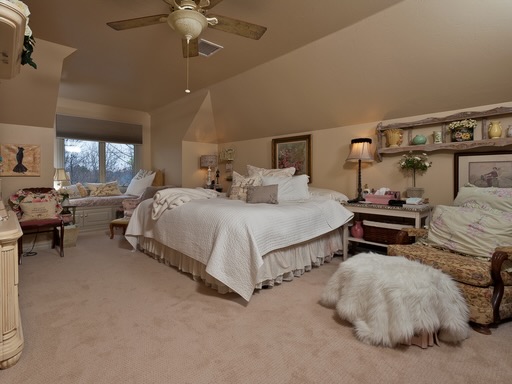}
&
\includegraphics[width=0.22\linewidth,height=15ex]{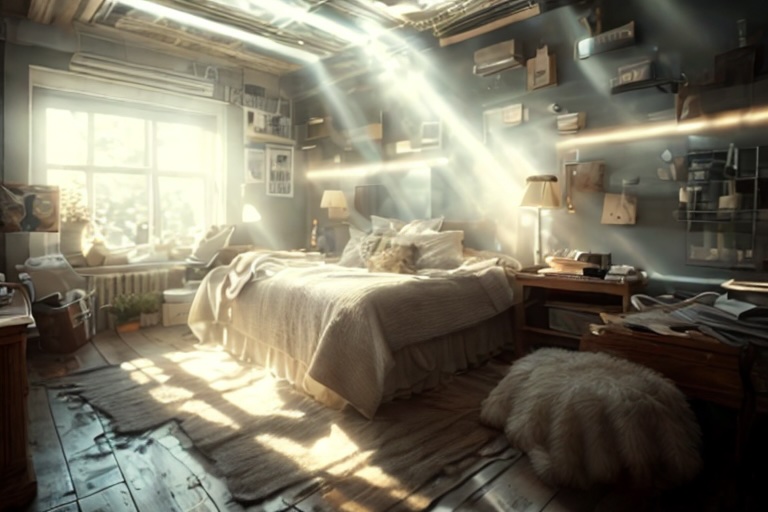} &
\includegraphics[width=0.22\linewidth,height=15ex]{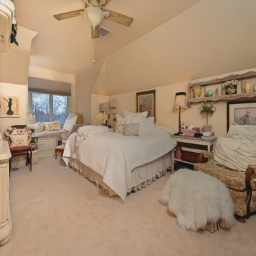} &
\includegraphics[width=0.22\linewidth,height=15ex]{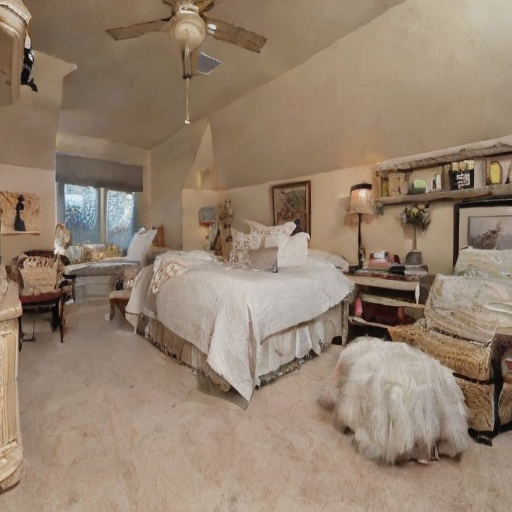} &
\includegraphics[width=0.22\linewidth,height=15ex]{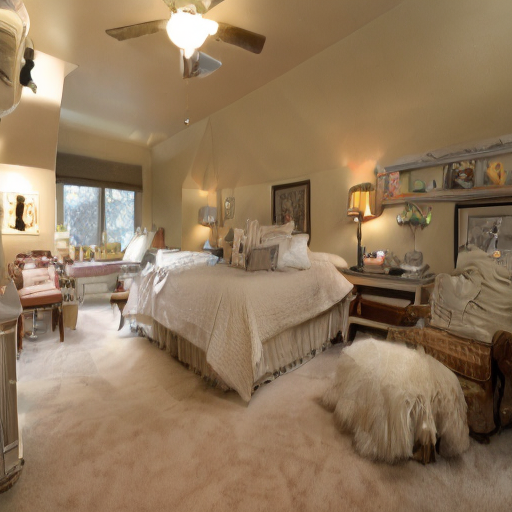} \\

\rotatebox{90}{\small Light 2} &
\includegraphics[width=0.22\linewidth,height=15ex]{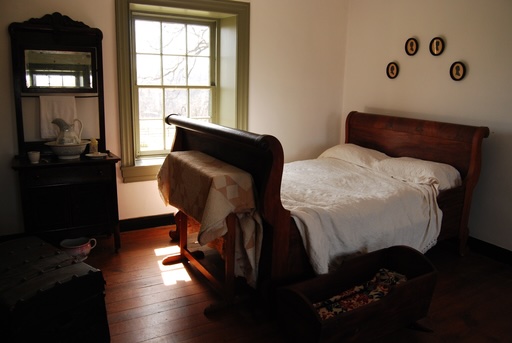} &
\includegraphics[width=0.22\linewidth,height=15ex]{figures/img_src/fig9_img/8100.jpg}
&
\includegraphics[width=0.22\linewidth,height=15ex]{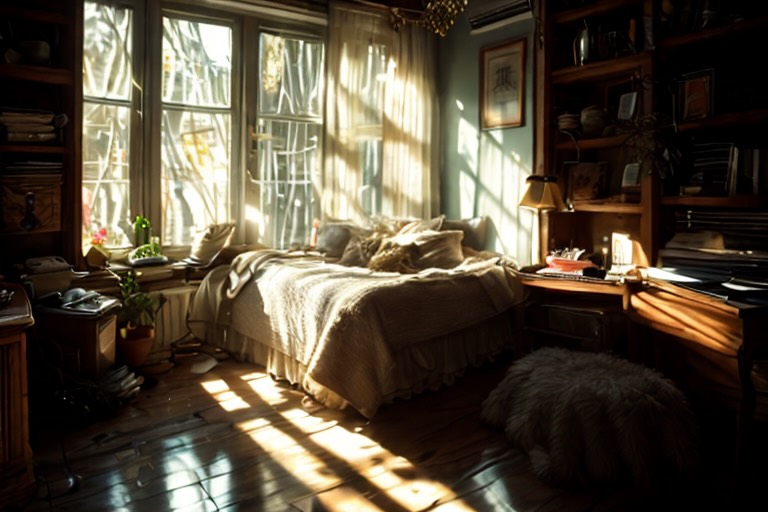} &
\includegraphics[width=0.22\linewidth,height=15ex]{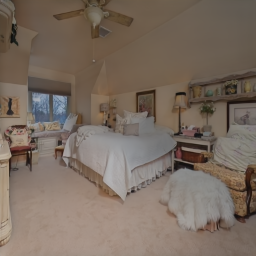}  &
\includegraphics[width=0.22\linewidth,height=15ex]{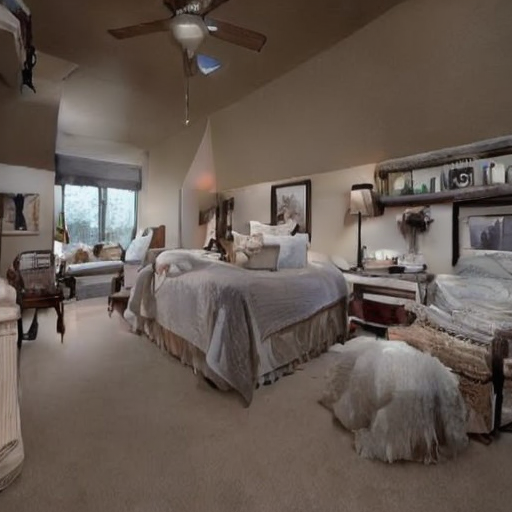} &
\includegraphics[width=0.22\linewidth,height=15ex]{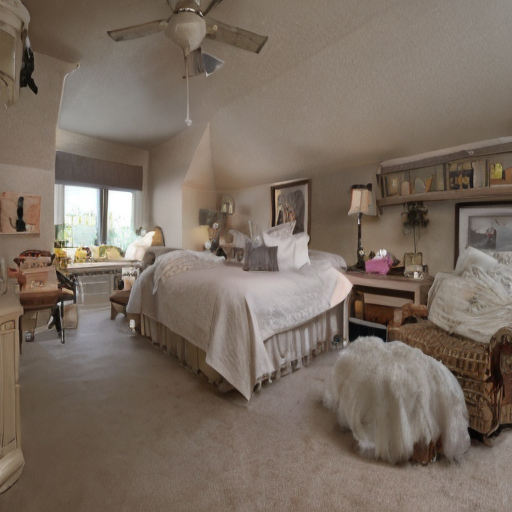} \\
\midrule
\rotatebox{90}{\small Light 1} &
\includegraphics[width=0.22\linewidth,height=15ex]{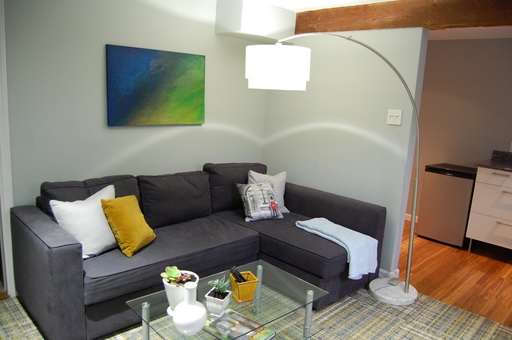} &
\includegraphics[width=0.22\linewidth,height=15ex]{figures/img_src/real_relight/91780.jpg}
&
\includegraphics[width=0.22\linewidth,height=15ex]{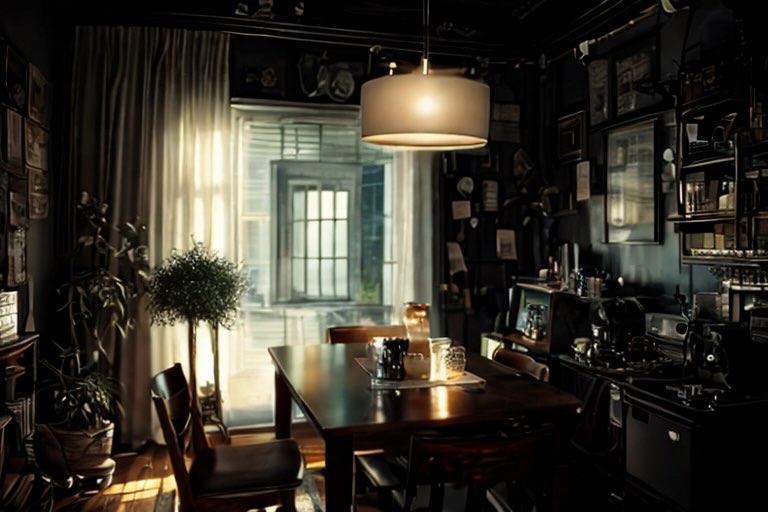} &
\includegraphics[width=0.22\linewidth,height=15ex]{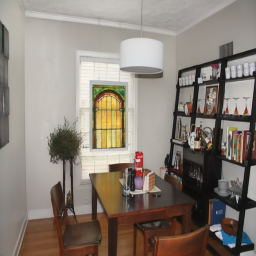}&
\includegraphics[width=0.22\linewidth,height=15ex]{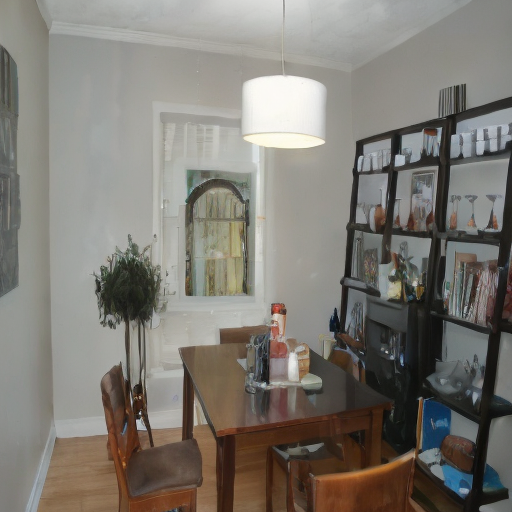} &
\includegraphics[width=0.22\linewidth,height=15ex]{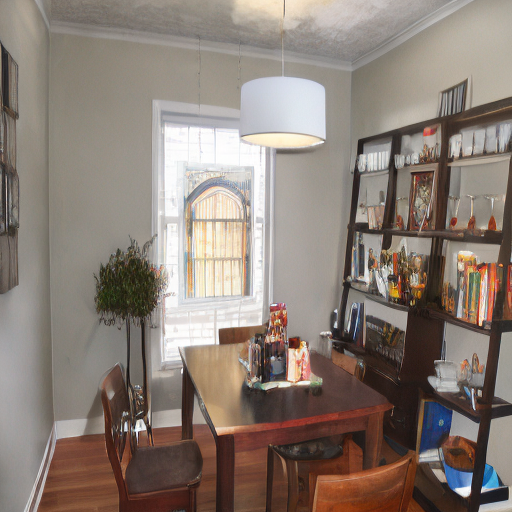} \\

\rotatebox{90}{\small Light 2} &
\includegraphics[width=0.22\linewidth,height=15ex]{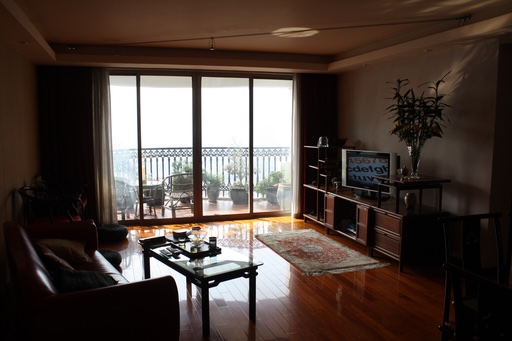} &
\includegraphics[width=0.22\linewidth,height=15ex]{figures/img_src/real_relight/91780.jpg}
&
\includegraphics[width=0.22\linewidth,height=15ex]{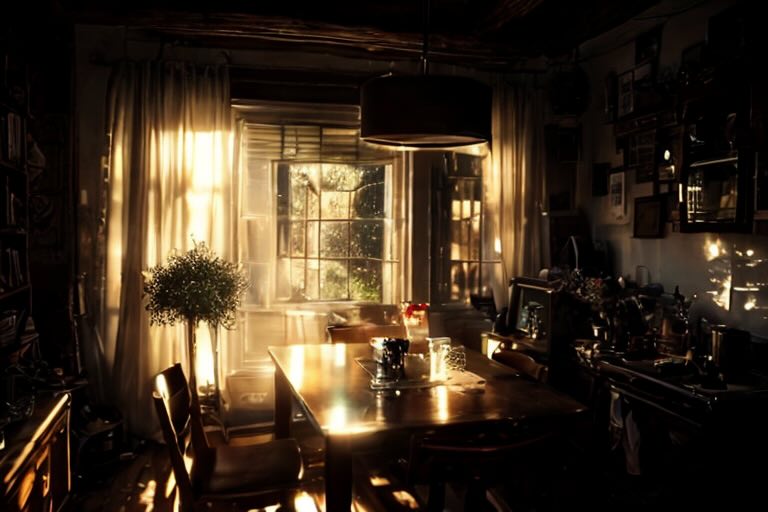} &
\includegraphics[width=0.22\linewidth,height=15ex] {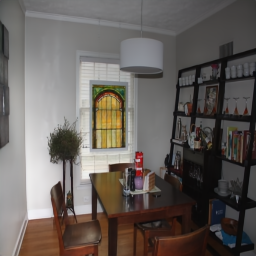} &
\includegraphics[width=0.22\linewidth,height=15ex] {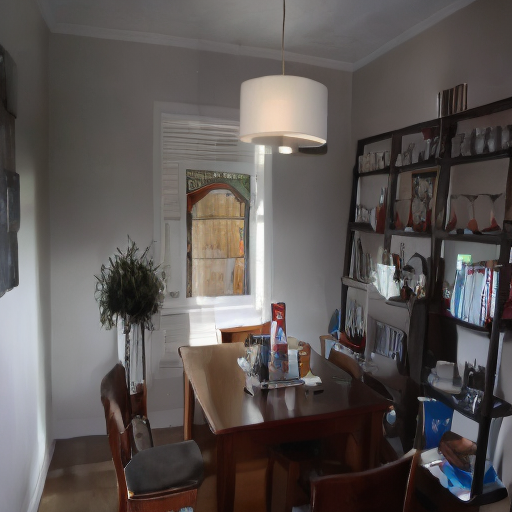} &
\includegraphics[width=0.22\linewidth,height=15ex]{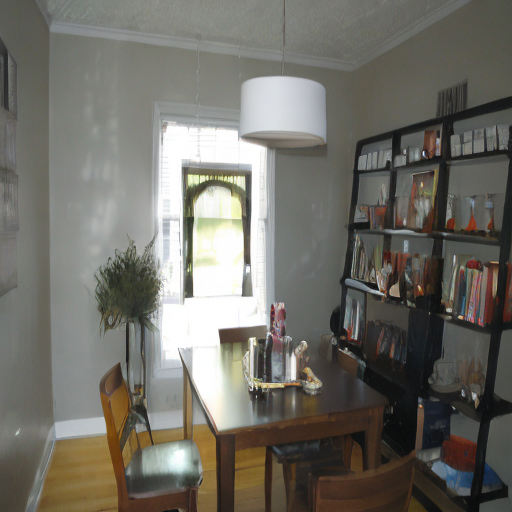} \\
 & \small Target Lighting & \small Original Image & \small IC-Light & \small Latent-intrinsics & \small LumiNet & \small Ours 
\end{tabular}}
\caption{\textbf{Relighting comparison across two real-world images}, each shown under two target illuminations (Light 1: lamp-dominated; Light 2: sunlight through windows). IC-Light~\citep{IC-Light} produces stylized results with exaggerated glow and artifacts that diverge from the targets. Latent-intrinsics~\citep{Latent-Intrinsic} captures some variation but yields low-contrast, flattened illumination with weak directionality. LumiNet~\citep{LumiNet} better matches global tone but remains overly diffuse, often missing dominant light sources and underestimating cast shadows and highlight localization. \textbf{Ours} preserves material detail and transfers both global and directional lighting, producing images that most closely match the targets lighting. }
\label{fig:real_m_relight_swapped}
\end{figure*}

\noindent\textbf{In-Scene Relighting.}
We also evaluate in-scene relighting on MIIW (Tab.~\ref{tab:MIIW-in-scene}), where each image captured under illumination $i$ is relit to the corresponding $(i+12)$ illumination condition. ALI achieves consistent performance and the second-best LPIPS without privileged labels. In contrast, UniRelight, despite using additional supervision, video-based training, and dataset-specific optimization, still exhibits noticeable failures in challenging material regions, as shown in Fig.~\ref{fig:miiw_comparison}.

\begin{table}[t]
  \centering
  \small
  \caption{
    \textbf{MIIW \textit{cross-scene} evaluation.} Following Latent-Intrinsics and LumiNet, we compare source and target images from different scenes and report RMSE and SSIM on raw and color-corrected outputs. Trained on diverse images without privileged labels, our method performs competitively—especially within its training category. \textbf{Bold} highlights the best in each block (MIIW-only vs.\ diverse images for training). Note that metrics favor tone consistency over shadow or specular accuracy; qualitative results remain the primary evidence of lighting fidelity~\citep{giroux2024towards}.
  }
  \label{tab:MIIW-cross}
  \begin{adjustbox}{max width=\linewidth}
  \begin{tabular}{@{}lc|cc|cc@{}}
    \toprule
    \cmidrule(lr){3-4}\cmidrule(l){5-6}
    \textbf{Methods} & \textbf{Labels}&RMSE$\downarrow$ &SSIM$\uparrow$ &RMSE$\downarrow$ &  SSIM$\uparrow$ \\
    \midrule
    \multicolumn{6}{l}{\textbf{Trained only on MIIW}}\vspace{2pt} \\
    SA\textendash AE~\citep{hu2020sa} & Light & {\textbf{{0.288}}} & {\textbf{{0.484}}} & 0.232 & 0.559 \\
    SA\textendash AE~\cite{hu2020sa} & --    & 0.443 & 0.300 & 0.317 & 0.431 \\
    S3Net~\citep{yang2021s3net}       & Depth & 0.512 & 0.331 & 0.418 & 0.374 \\
    S3Net~\citep{yang2021s3net}       & --    & 0.499 & 0.336 & 0.414 & 0.377 \\
    Latent\textendash Intrinsic~\citep{Latent-Intrinsic} & -- & {0.297} & {0.473} & \textbf{0.222} & \textbf{0.571} \\
    \midrule
    \multicolumn{6}{l}{\textbf{Trained on diverse indoor images}}\vspace{2pt} \\
    RGB$\leftrightarrow$X~\citep{zeng2024rgb} & G-Buffer & 0.587 & 0.070 & 0.427 & 0.215 \\
    DiffusionRenderer~\citep{DiffusionRenderer2025} & Env.\ map & 0.399 & 0.354 & 0.341 & 0.355 \\
    LumiNet~\citep{LumiNet} & -- & 0.310 & 0.440 & 0.240 & 0.527 \\
    \textbf{Ours} & -- & \textbf{0.294} & \textbf{0.464} & \textbf{0.231} & \textbf{0.553} \\
    \bottomrule
  \end{tabular}
  \end{adjustbox}
\end{table}

\begin{table}[t]
  \centering
  \small
  \caption{\textbf{MIIW \textit{in-scene} evaluation.}
  Results on the multi-illumination dataset of~\cite{MIIW}, where source and target are the same scene captured under
  different illuminations. We report PSNR, RMSE, LPIPS, and SSIM. Our method, trained on diverse images without privileged
  labels, achieves competitive results compared to prior approaches. Higher PSNR/SSIM and lower RMSE/LPIPS indicate better
  performance. \textsuperscript{\dag} trained only on MIIW; \textsuperscript{*} numbers reported from UniRelight.
  Pixel metrics can be sensitive to small misalignments~\citep{giroux2024towards}.}
  \label{tab:MIIW-in-scene}
  \begin{adjustbox}{max width=\linewidth}
  \begin{tabular}{@{}lc|cccc@{}}
    \toprule
    \textbf{Methods} & \textbf{Labels} & PSNR$\uparrow$ & RMSE$\downarrow$ & LPIPS$\downarrow$ & SSIM$\uparrow$ \\
    \midrule
    RGB$\leftrightarrow$X~\citep{zeng2024rgb} & G-Buffer     & 15.674   & 0.161   & 0.323   & 0.500   \\
    DiffusionRenderer~\citep{DiffusionRenderer2025} & Env.\ map & 16.810   & 0.156   & 0.343   & 0.612   \\
    UniRelight~\citep{he2025unirelight}\textsuperscript{*}
                                                  & Albedo, Env.\ map & {20.760} & --    & 0.251 & {0.749} \\
    \midrule
    Latent-Intrinsics~\citep{Latent-Intrinsic}\textsuperscript{\dag} & -- & \textbf{21.350} & \textbf{0.092} & \textbf{0.157} & \textbf{0.794} \\
    LumiNet~\citep{LumiNet} & -- & 18.568 & 0.123 & 0.228 & 0.645 \\
    \textbf{Ours}                                  & --        & 18.872 & {0.119} & {0.213} & 0.671 \\
    \bottomrule
  \end{tabular}
  \end{adjustbox}
\end{table}

\begin{table}[ht]
\centering
\caption{
\textbf{User Study.}
We conduct two pairwise user studies: (a) an in-the-wild comparison among LumiNet, Latent-Intrinsics, and our method; and (b) a stage-wise evaluation across our training pipeline. Users are shown two relit images and asked to select the one with better lighting alignment, identity preservation, and lighting realism. Latent-Intrinsics better preserves identity, but struggles with lighting alignment and realism. Our method achieves stronger lighting realism and alignment, with Stage III further improving perceptual quality. $^{*}\,p<0.05$
}
\label{tab:usr_study}
\footnotesize
\setlength{\tabcolsep}{3.5pt}
\textbf{(a) In-the-wild relighting} $\uparrow$
\vspace{2pt}
\resizebox{0.75\linewidth}{!}{
\begin{tabular}{lccc}
\toprule
& Latent-Intrinsics & LumiNet & Ours \\
\midrule
Lighting alignment    & 0.13$^{*}$ & 0.42 & \textbf{0.93}$^{*}$ \\
Identity preservation & \textbf{0.68}$^{*}$ & 0.05$^{*}$ & 0.63$^{*}$ \\
Lighting realism      & 0.28$^{*}$ & 0.75$^{*}$ & \textbf{0.90}$^{*}$ \\
\bottomrule
\end{tabular}
}
\vspace{8pt}

\textbf{(b) Stage-wise evaluation} $\uparrow$
\vspace{2pt}
\resizebox{0.8\linewidth}{!}{
\begin{tabular}{lcccc}
\toprule
& LumiNet & Stage I & Stage I\&II & All Stages \\
\midrule
Lighting alignment    & 0.44 & 0.20$^{*}$ & 0.21$^{*}$ & \textbf{0.75}$^{*}$ \\
Identity preservation & 0.05$^{*}$ & 0.45 & 0.68$^{*}$ & \textbf{0.96}$^{*}$ \\
Lighting realism      & 0.25$^{*}$ & 0.35$^{*}$ & 0.68$^{*}$ & \textbf{0.89}$^{*}$ \\
\bottomrule
\end{tabular}
}
\vspace{-1ex}
\end{table}

\subsection{Relighting as a Probe of Visual Priors}
\noindent\textbf{Which Visual Priors Are Relightable?}
We next use ALI as a controlled probe to study how different pretrained visual encoders affect relighting. We keep the relighting backbone, training data, and decoder objective fixed, and vary only the frozen encoder used for semantic augmentation. This isolates the contribution of each visual prior to relightability.

Table~\ref{tab:semantic-impact} compares CLIP, DINOv2/3, MAE, and RADIOv2.5. MAE and RADIOv2.5 consistently outperform CLIP and DINO-based features. This suggests that semantic recognition strength alone is not sufficient for relighting. Contrastive and distillation-based encoders often promote invariance to lighting, texture, and local appearance, which are precisely the signals required for light transfer. In contrast, MAE's pixel-reconstruction objective better preserves fine-grained color, shading, and reflectance cues. RADIOv2.5 further improves this balance by providing dense, pixel-aligned features distilled from complementary teachers. These results support our central finding: relightability depends on balancing semantic context with photometric fidelity.

This observation is consistent with concurrent work RAE~\citep{zheng2025diffusion}, which finds that an MAE-based representation paired with a diffusion decoder outperforms a DINO-based counterpart for image reconstruction under the same decoding architecture. Our results extend this insight to relighting, where preserving photometric structure is not only useful for reconstruction but essential for physically meaningful illumination transfer.

\begin{table}[t]
  \centering
  \footnotesize
  \caption{\textbf{Impact of semantic features on relighting.}
  Augmenting latent intrinsics with RADIOv2.5 or MAE significantly improves metrics after Stage~II.}
  \label{tab:semantic-impact}
  \begin{adjustbox}{max width=\linewidth}
  \begin{tabular}{@{}lccccc@{}}
    \toprule
    \textbf{Feature} & \textbf{Stage} & \textbf{RMSE} $\downarrow$ & \textbf{LPIPS} $\downarrow$ & \textbf{PSNR} $\uparrow$ & \textbf{SSIM} $\uparrow$ \\
    \midrule
    Latent Intrinsic~\cite{Latent-Intrinsic} & I       & 0.1380 & 0.2857 & 17.5763 & 0.5461 \\
                                             & I\&II   & 0.1383 & 0.2844 & 17.5463 & 0.5531 \\
    \midrule
    MAE~\cite{he2022masked_mae}             & I       & 0.2195 & 0.4820 & 14.2381 & 0.4571 \\
                                             & I\&II   & 0.1286 & 0.2554 & 17.9861 & 0.4852 \\
    \midrule
    DINOv2~\cite{caron2021emerging}         & I       & 0.2786 & 0.3295 & 13.2439 & 0.4646 \\
                                             & I\&II   & 0.1686 & 0.3253 & 15.7945 & 0.4815 \\
    \midrule
    DINOv3~\cite{simeoni2025dinov3}         & I       & 0.1794 & 0.3588 & 15.1375 & 0.4824 \\
                                             & I\&II   & 0.1654 & 0.2923 & 16.1510 & 0.5299 \\
    \midrule
    CLIP~\cite{radford2021learning}         & I       & 0.2189 & 0.4556 & 13.9671 & 0.3987 \\
                                             & I\&II   & 0.1627 & 0.3153 & 16.1333 & 0.5039 \\
    \midrule
    RADIOv2.5~\cite{heinrich2025radiov25improvedbaselinesagglomerative} & I     & 0.1312 & 0.2673 & 17.9448 & 0.5609 \\
                                                                         & I\&II & \textbf{0.1260} & \textbf{0.2440} & \textbf{18.3426} & \textbf{0.5958} \\
    \bottomrule
  \end{tabular}
  \end{adjustbox}
  \vspace{-10pt}
\end{table}

\noindent\textbf{Material-Wise Relightability.}
To further probe representation quality, we evaluate relighting performance within material-specific regions. We group per-pixel MIIW material labels into five physically motivated categories: \textit{Diffuse}, \textit{Glossy}, \textit{Specular}, \textit{Metallic}, and \textit{Uncertain/Mixed}. We then compute SSIM, PSNR, and RMSE within each region.

As shown in Tab.~\ref{tab:metarial_metrics}, semantic augmentation in Stage~I improves performance across all material categories, with the largest gains on non-diffuse materials such as \textit{Glossy}, \textit{Metallic}, and \textit{Specular}. These regions require semantic context to distinguish material appearance from illumination effects, while also requiring dense photometric cues to reproduce highlights and view-dependent effects. Because specular and translucent regions occupy only a small fraction of the semantic masks, region-averaged metrics likely understate the perceptual improvement on the most challenging pixels.

\begin{table*}[ht]
\caption{\textbf{Performance comparison across different material semantic labels.} Background color represents relative improvement within each metric, redder (warmer) shades indicating greater enhancement compared to the baseline LumiNet, which utilizes latent intrinsic features as intrinsic representation. Large gains are observed in complex metallic, glossy or specular surfaces. Note because scene/protocol/metric are fixed for this evaluation, PSNR/SSIM/RMSE serve as internal proxies: the monotonic improvements from Stage I→II→All-Stage confirm better photometric calibration without regressions; perceptual lighting gains are corroborated by the visual comparisons and directional cues.}
\label{tab:metarial_metrics}
\centering
\resizebox{1\textwidth}{!}{

\begin{tabular}{l|ccc|ccc|ccc|ccc}
\toprule
\textbf{Semantic Label} & \multicolumn{3}{c|}{\textbf{LumiNet} \cite{LumiNet}} & \multicolumn{3}{c|}{\textbf{Ours (Stage I Only)}} & \multicolumn{3}{c}{\textbf{Ours (Stage I\&II)}} & \multicolumn{3}{c}{\textbf{Ours (All Stage)}} \\
& SSIM↑ & PSNR↑ & RMSE↓ & SSIM↑ & PSNR↑ & RMSE↓ & SSIM↑ & PSNR↑ & RMSE↓ & SSIM↑ & PSNR↑ & RMSE↓\\
\midrule

Uncertain&\cellcolor[rgb]{1.0,1.0,0.7}0.6082&\cellcolor[rgb]{1.0,1.0,0.7}17.8499&\cellcolor[rgb]{1.0,1.0,0.7}0.1426&\cellcolor[rgb]{1.0,0.85,0.45}0.6178&\cellcolor[rgb]{1.0,0.9,0.5}18.0678&\cellcolor[rgb]{1.0,0.82,0.4}0.1373&\cellcolor[rgb]{0.98,0.5,0.25}0.6518&\cellcolor[rgb]{0.98,0.6,0.28}18.6282&\cellcolor[rgb]{0.98,0.62,0.28}0.1307&\cellcolor[rgb]{0.98,0.55,0.26}0.6460&\cellcolor[rgb]{0.98,0.68,0.32}18.4975&\cellcolor[rgb]{0.98,0.63,0.28}0.1310\\

Diffuse&\cellcolor[rgb]{1.0,1.0,0.7}0.5281&\cellcolor[rgb]{1.0,1.0,0.7}18.0883&\cellcolor[rgb]{1.0,1.0,0.7}0.1369&\cellcolor[rgb]{1.0,0.88,0.48}0.5464&\cellcolor[rgb]{1.0,0.86,0.48}18.3033&\cellcolor[rgb]{1.0,0.8,0.42}0.1320&\cellcolor[rgb]{0.98,0.62,0.28}0.5798&\cellcolor[rgb]{0.98,0.58,0.26}18.7476&\cellcolor[rgb]{0.98,0.56,0.25}0.1261&\cellcolor[rgb]{0.98,0.68,0.32}0.5731&\cellcolor[rgb]{0.98,0.64,0.3}18.6726&\cellcolor[rgb]{0.98,0.6,0.27}0.1269\\

Glossy&\cellcolor[rgb]{1.0,1.0,0.7}0.5720&\cellcolor[rgb]{1.0,1.0,0.7}18.6291&\cellcolor[rgb]{1.0,1.0,0.7}0.1296&\cellcolor[rgb]{1.0,0.8,0.42}0.5988&\cellcolor[rgb]{1.0,0.78,0.45}19.0248&\cellcolor[rgb]{1.0,0.7,0.38}0.1217&\cellcolor[rgb]{0.98,0.55,0.26}0.6291&\cellcolor[rgb]{0.95,0.45,0.22}19.6906&\cellcolor[rgb]{0.95,0.45,0.22}0.1152&\cellcolor[rgb]{0.98,0.62,0.3}0.6196&\cellcolor[rgb]{0.96,0.52,0.26}19.5275&\cellcolor[rgb]{0.95,0.47,0.23}0.1157\\

Metallic&\cellcolor[rgb]{1.0,1.0,0.7}0.4164&\cellcolor[rgb]{1.0,1.0,0.7}15.4926&\cellcolor[rgb]{1.0,1.0,0.7}0.1759&\cellcolor[rgb]{1.0,0.96,0.55}0.4608&\cellcolor[rgb]{1.0,0.96,0.52}16.0831&\cellcolor[rgb]{1.0,0.95,0.5}0.1636&\cellcolor[rgb]{1.0,0.92,0.4}0.4855&\cellcolor[rgb]{1.0,0.93,0.42}16.3285&\cellcolor[rgb]{1.0,0.93,0.4}0.1588&\cellcolor[rgb]{1.0,0.92,0.4}0.4822&\cellcolor[rgb]{1.0,0.92,0.38}16.3827&\cellcolor[rgb]{1.0,0.92,0.38}0.1581\\

Specular&\cellcolor[rgb]{1.0,1.0,0.7}0.3778&\cellcolor[rgb]{1.0,1.0,0.7}17.0860&\cellcolor[rgb]{1.0,1.0,0.7}0.1490&\cellcolor[rgb]{1.0,0.95,0.58}0.4120&\cellcolor[rgb]{1.0,0.8,0.42}17.6720&\cellcolor[rgb]{1.0,0.78,0.4}0.1394&\cellcolor[rgb]{1.0,0.9,0.38}0.4423&\cellcolor[rgb]{0.98,0.72,0.34}17.9798&\cellcolor[rgb]{0.98,0.74,0.36}0.1357&\cellcolor[rgb]{1.0,0.9,0.38}0.4365&\cellcolor[rgb]{0.98,0.68,0.3}18.0688&\cellcolor[rgb]{0.98,0.65,0.28}0.1330\\
\bottomrule
\end{tabular}
}
\end{table*}

\noindent\textbf{Intrinsic--Extrinsic Disentanglement.}
ALI learns a representation that separates scene-invariant intrinsic structure from illumination-dependent extrinsic codes. Figures~\ref{fig:mit_m_relight} and~\ref{fig:real_m_relight_swapped} show that holding intrinsics fixed while varying the lighting input produces coherent relighting. Moreover, Fig.~\ref{fig:interp_combined} shows that interpolating the extrinsic code yields smooth and physically plausible lighting transitions within the same scene. This supports the interpretation that ALI improves relightability by augmenting latent intrinsics without collapsing lighting and content information.
\begin{figure*}[t]
\centering
\begin{subfigure}{\linewidth}
    \includegraphics[width=\linewidth]{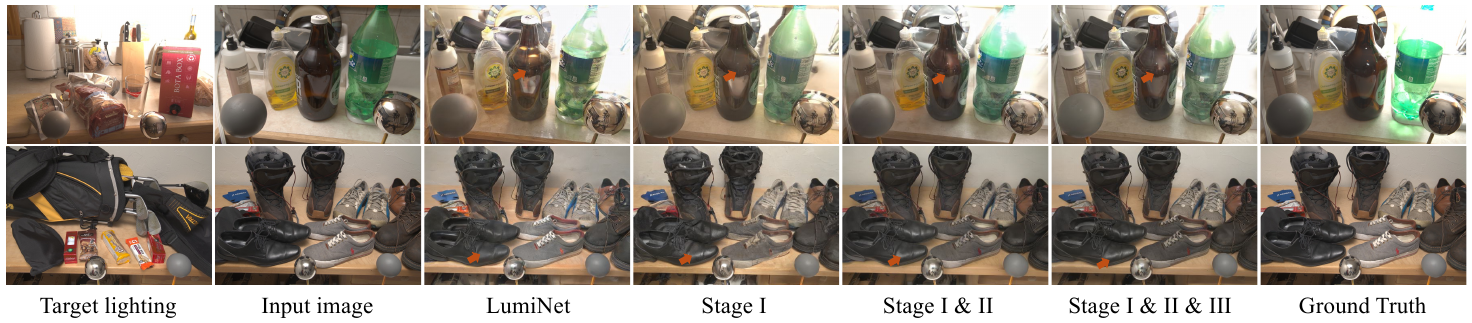}
    \caption{Multi-stage ablation on MIIW \cite{MIIW} dataset.}
\end{subfigure}
\\\vspace{5pt}
\begin{subfigure}{\linewidth}
    \includegraphics[width=\linewidth]{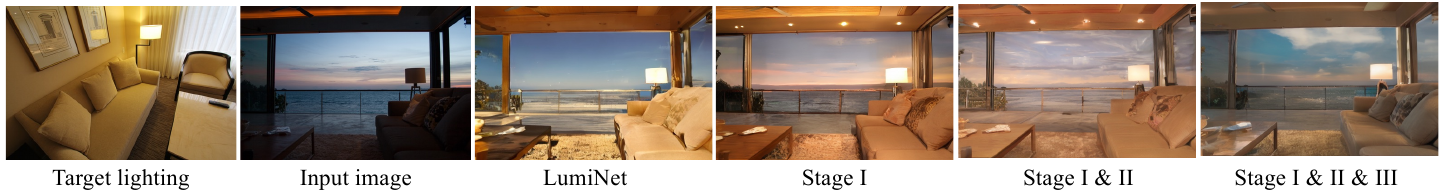}
    \caption{Multi-stage ablation on in-the-wild image (with bypass decoder disabled).}
\end{subfigure}
\vspace{-5pt}
\caption{\textbf{Multi-stage ablation.} \textbf{\textit{Top}}: Compared to LumiNet, our Stage I improves fine geometry details. Adding Stage II sharpens directional cues and specular effects, while the full pipeline (Stage I\&II\&III) produces the closest match to ground truth, with accurate shadows, highlights, and material fidelity. This progression illustrates how each stage contributes complementary improvements, consistent with the quantitative gains in Tab.~\ref{tab:metarial_metrics}.  \textbf{\textit{Bottom:}} LumiNet produces flat illumination with weak lamp cues. Stage I introduces coarse global tone but has color shift, Stage II suppresses these effects, and the full pipeline (Stage I\&II\&III) yields the most faithful transfer: interior warmth from the lamp is preserved while maintaining the outdoor scene, closely matching the target lighting. This demonstrates that our stage-wise design generalizes to unconstrained real-world images.}
\label{fig:metal_ablation}
\vspace{-2ex}
\end{figure*}

\begin{figure*}[htpb!]
    \centering
    \begin{subfigure}{\linewidth}
        \centering
        \includegraphics[width=\linewidth]{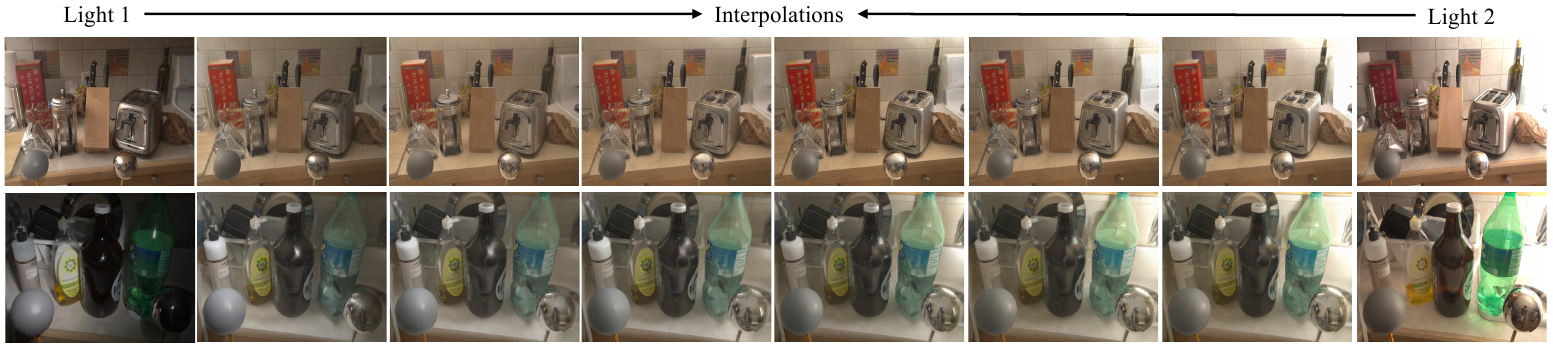}
        \caption{Image relighting with interpolated lighting code. }
        \label{fig:interp_a}
    \end{subfigure}
    \\\vspace{5pt} %
    \begin{subfigure}{\linewidth}
        \centering
        \includegraphics[width=\linewidth]{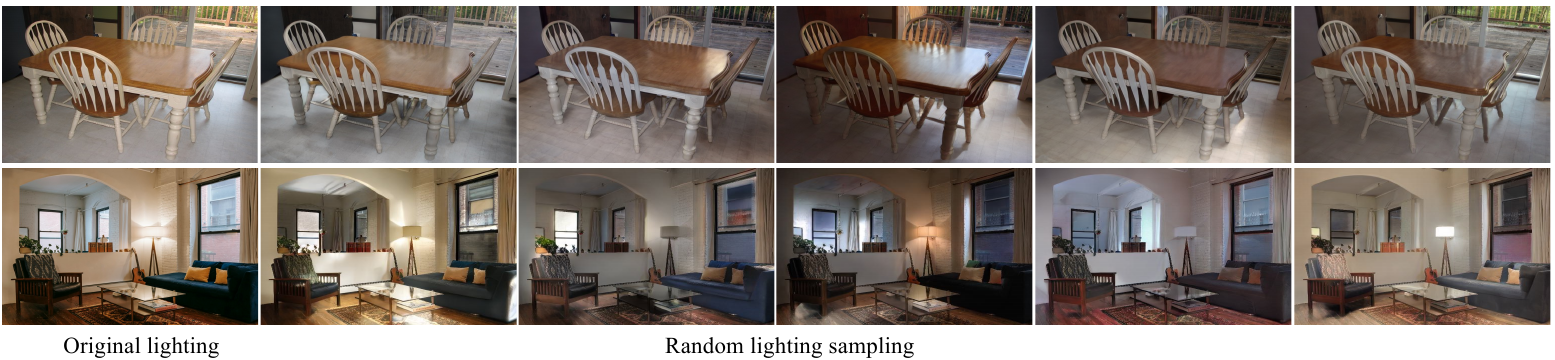}
        \caption{Zero-Shot image relighting with unpaired, randomly sampled lighting code.}
        \label{fig:interp_b}
    \end{subfigure}
    \vspace{-5pt}
    \caption{\textbf{Lighting interpolation and diversity.} \textbf{Top:} Generated images showing a smooth interpolation between two lighting codes. Note the plausible evolution of directional lighting, including the progressive appearance of sharp specular highlights on the toaster and caustic effects from the bottle.
\textbf{Bottom:} In-the-wild relighting results using lighting codes sampled from random, unpaired images. Our method produces a diverse range of distinct illumination effects, plausible altering the glossy reflections on the dining table and the ambient lighting in the living room.
}
    \label{fig:interp_combined}
    \vspace{-10pt}
\end{figure*}

\section{Discussion}

This work studies image-to-image relighting as a probe of physically grounded visual priors. Our results reveal a counterintuitive finding: features from strong semantic encoders can degrade, rather than improve, high-fidelity relighting. This suggests that recognition-oriented representations, while powerful for semantic abstraction, often suppress the illumination-sensitive details needed for physical image transformation. Relighting therefore exposes a semantic--photometric trade-off: useful features must preserve dense appearance cues while still providing enough semantic context to resolve material and object-level ambiguities.

\textbf{Relightability depends on representation bias, not semantic strength alone.}
Our backbone analysis shows that stronger semantic encoders are not necessarily more relightable. Increasing feature density, as in DINOv2 to DINOv3, is insufficient when the underlying objective promotes invariance to texture, lighting, and local appearance. In contrast, MAE-style reconstruction objectives better preserve pixel-aligned structure and photometric detail, which are essential for light transfer. RADIOv2.5 further improves this balance by combining dense spatial alignment with complementary semantic cues from multi-teacher distillation. These results suggest that relightability depends less on a specific backbone family than on the inductive biases imposed by reconstruction, spatial alignment, and feature granularity.

\textbf{Prior design can substitute for supervision scale.}
ALI belongs to a broader class of prior-driven generative methods. Rather than relying on large synthetic datasets, dense physical annotations, or explicit inverse-graphics supervision, ALI injects a frozen visual prior into a latent-intrinsic framework. This lightweight fusion improves both material fidelity and relighting robustness, especially for glossy and specular surfaces. In contrast to scale-centric approaches that attempt to learn physical structure from large supervised datasets~\citep{zeng2024rgb, DiffusionRenderer2025, he2025unirelight}, our results indicate that careful prior selection and integration can provide a more data-efficient route to physically grounded generation.

\textbf{Limitations and future work.}
ALI improves material-aware relighting but does not solve inverse rendering. It can still shift fine details, conflate surface color with illumination, and hallucinate shadows from learned priors rather than explicit geometry. These limitations are inherent to image-only relighting without 3D structure or physical supervision. Extending this probing framework to related tasks, such as reflectance editing, shadow manipulation, view-consistent generation, or material decomposition, may further clarify which representation properties support physical reasoning in generative models.

\section*{Impact Statement}

This work contributes to the understanding of how visual representations support physically grounded image generation, using relighting as a diagnostic task. By showing that stronger semantic encoders can degrade relighting performance, our findings may influence how pretrained visual priors are selected and integrated into generative models for graphics and computational photography applications. Improved relighting models can benefit content creation, augmented reality, and visual effects by enabling more realistic lighting edits without requiring synthetic data or privileged supervision. We do not anticipate direct negative societal impacts from this work; however, as with other image generation technologies, downstream applications should consider ethical use and potential misuse of generated imagery.

\bibliography{main,bibs/bib_mani,bibs/bigdaf,bibs/dafjulia,bibs/dafsupp,bibs/jasonrefs,bibs/lana,bibs/references,bibs/refs,bibs/in_GS,bibs/retinexdiff}
\bibliographystyle{icml2026}

\newpage
\appendix
\onecolumn

\appendix
\renewcommand{\thetable}{S.\arabic{table}}
\renewcommand{\thefigure}{S.\arabic{figure}}
\setcounter{table}{0}
\setcounter{figure}{0}

\section{Implementation details} 
\subsection{Training}
All experiments are conducted on a node equipped with 8 NVIDIA A6000 Ada 48GB GPUs. The model is trained at a resolution of $512 \times 512$ with an effective batch size of 64 (including gradient accumulation). 

For Stage I, we employ a scene-aware sampler that ensures each batch contains images from the same scene captured under different lighting conditions. For Stage II and Stage III, we use standard random sampling across scenes. Additionally, in Stage III, we include a small proportion {(10\% of the training set)} of identity relighting samples—where the input and target share the same lighting—to explicitly enforce geometry preservation.

Stage I is trained for 4 epochs, while Stage II and Stage III are trained for 2 epochs each. Due to differences in model size and parameter updates, Stage I training takes approximately 8 hours on a single GPU, while each of Stage II and III takes around 10 hours. We use the AdamW optimizer for all training stages.
\vspace{-2ex}
\subsection{Lighting zoo generation}
\label{sec:lightzoo}
\begin{figure}
    \centering
    \includegraphics[width=\linewidth]{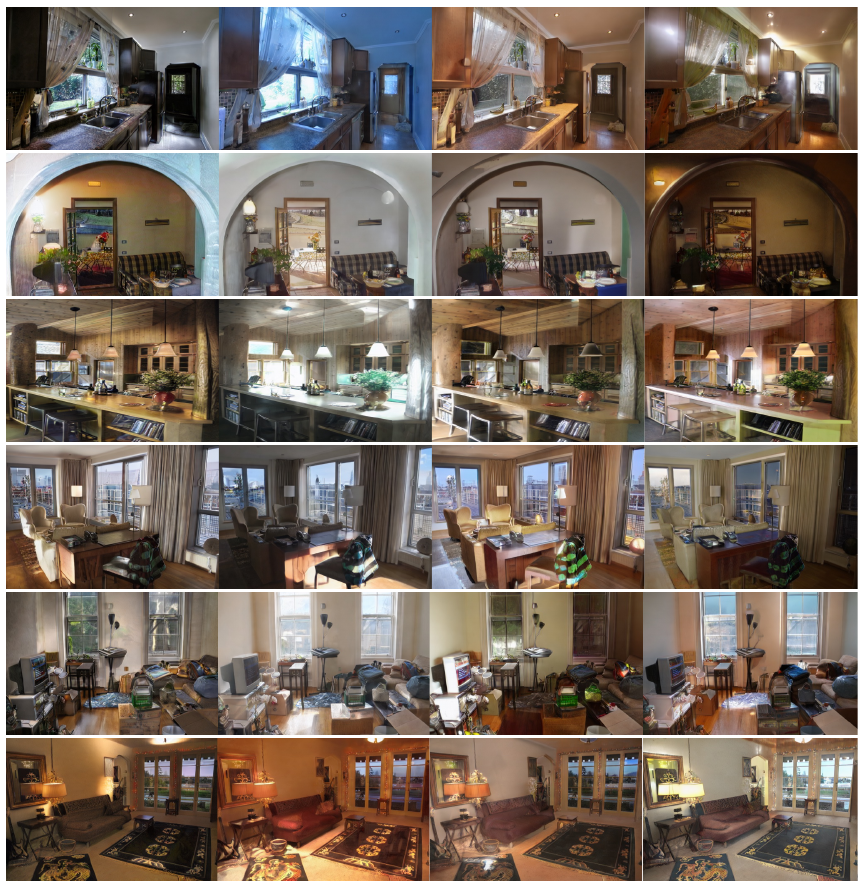}
    \caption{\textbf{Lighting Zoo.} Pseudo-relit image pairs generated by our method for the third-stage refinement.}
    \label{fig:Lightzoo}
\end{figure}

To enhance the model's generalization to real-world inputs, we use the Stage II checkpoint to generate a large-scale in-the-wild relighting dataset, referred to as the \textit{Lighting Zoo}. We curate approximately 6,000 diverse images from open-source datasets including IIW \citep{bell14intrinsic}, RealEstate-10K \citep{zhou2018stereo}, and DL3DV \citep{ling2024dl3dv}. Fig.~\ref{fig:Lightzoo} shows a selection of pseudo-relit pairs; note the plausibility and diversity of the synthesized lighting.

During generation, images are randomly grouped into batches. Each original image serves as the input, while the target lighting condition is randomly sampled from other images within the same batch. For each scene, we generate seven relit versions under different lighting conditions. The process is distributed across multiple NVIDIA A6000 Ada 48GB GPUs for efficiency. In total, generating the Lighting Zoo requires approximately 500 GPU hours.

\vspace{-2ex}
\subsection{User study}
We conduct two user studies: (1) comparing real-world performance of LumiNet, Latent Intrinsics, and our method; and (2) a stepwise assessment of our training pipeline's phases. Participants view two relit images—from different methods (Study 1) or different pipeline stages (Study 2)—and select the one with superior lighting alignment, identity preservation, and lighting realism. 23 participants completed all comparisons. Study 1 had each participant judge 9 image pairs across the three methods; Study 2 had each judge 12 pairs across four stages (LumiNet, Stage I, Stage I\&II, All Stages). All pairs were shown side-by-side at full resolution in randomized order with win/tie/lose selection.

\section{Semantic alignment in relighting}
\subsection{Semantic group curation}

The MIIW dataset~\cite{MIIW} provides detailed semantic annotations, including binary masks for 31 distinct material labels. To efficiently evaluate relighting performance across materials with varying reflectance properties, we regroup these fine-grained labels into five broader categories: \textit{Diffuse}, \textit{Glossy}, \textit{Specular}, \textit{Metallic}, and \textit{Uncertain/Mixed}. 

To ensure consistency and physical relevance, we utilize advanced reasoning model GPT-O3 with prompt - "cluster those class label into a higher-level BRDF-style categories." - to assist in the semantic reasoning and grouping process. The full mapping between original material labels and grouped categories is provided in Table~\ref{tab:mat_class}.

\section{More Visual Results}
\begin{figure*}[t]
\centering
\renewcommand{\arraystretch}{0.4}
\setlength{\tabcolsep}{1pt}
\resizebox{\linewidth}{!}{
\begin{tabular}{cccccc}
\includegraphics[width=0.16\linewidth]{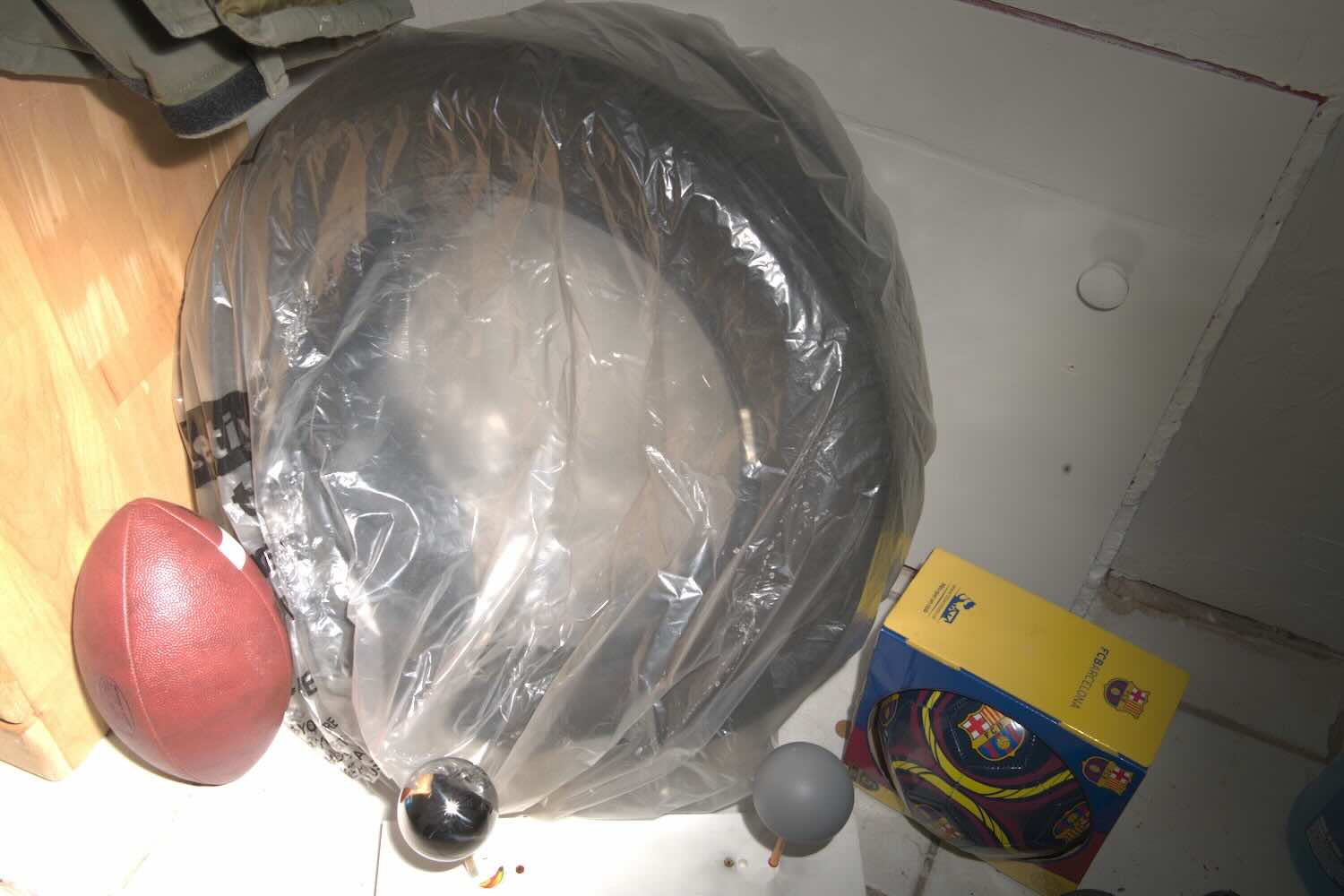} &
\includegraphics[width=0.16\linewidth]{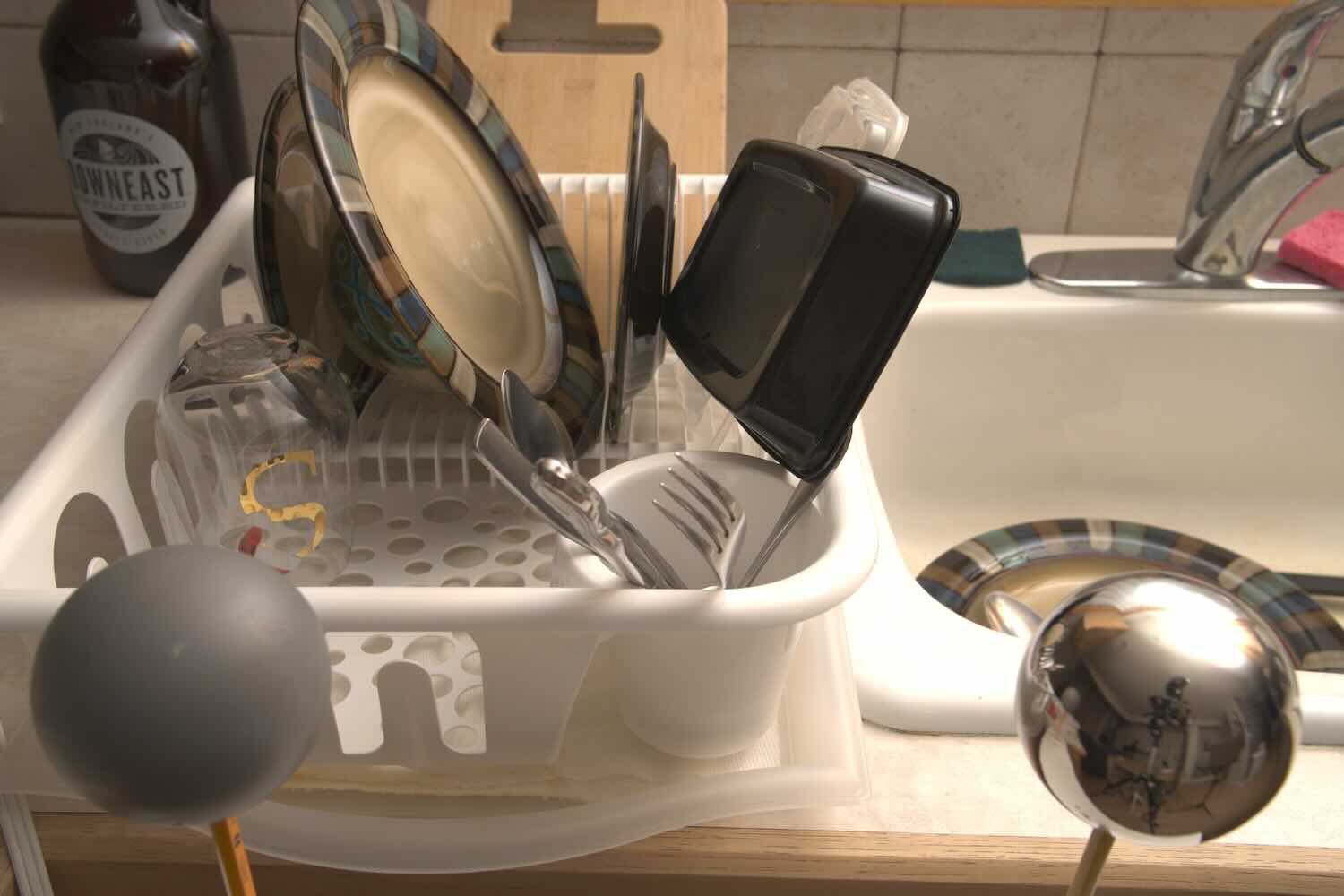} &
\includegraphics[width=0.16\linewidth]{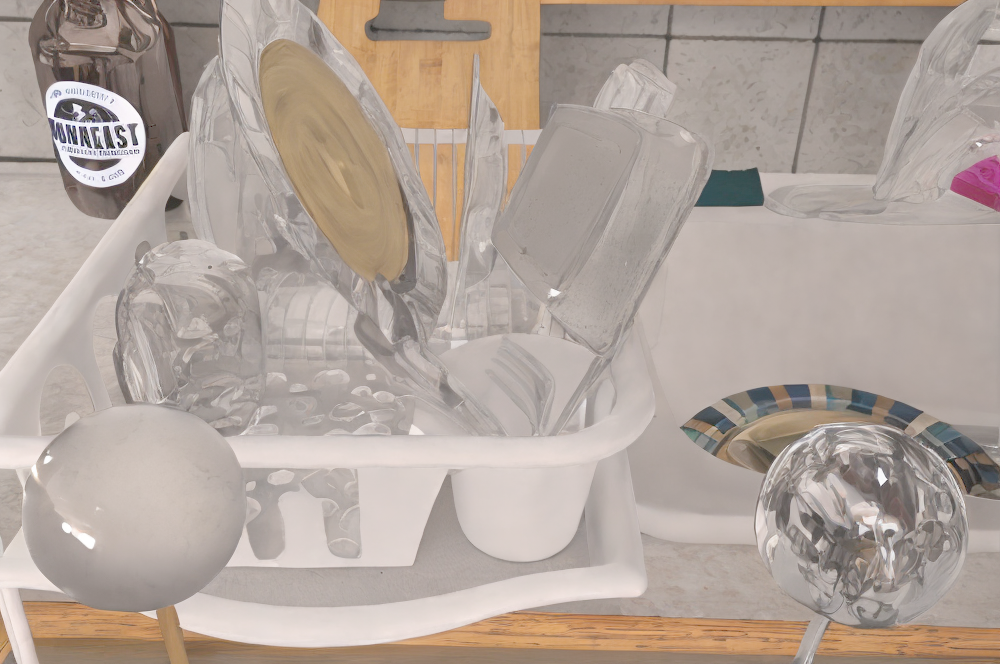} &
\includegraphics[width=0.16\linewidth,height=11.53ex]{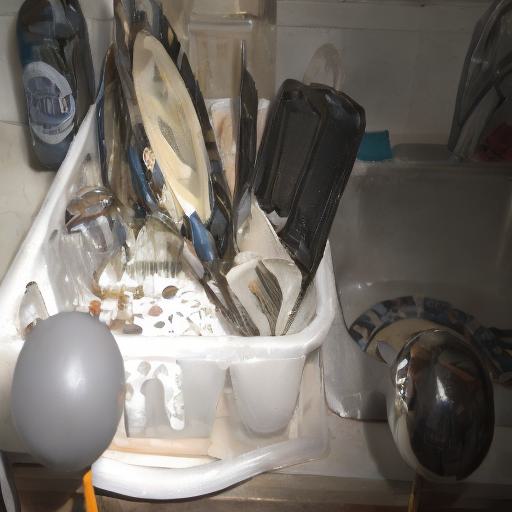} &
\includegraphics[width=0.16\linewidth,height=11.53ex]{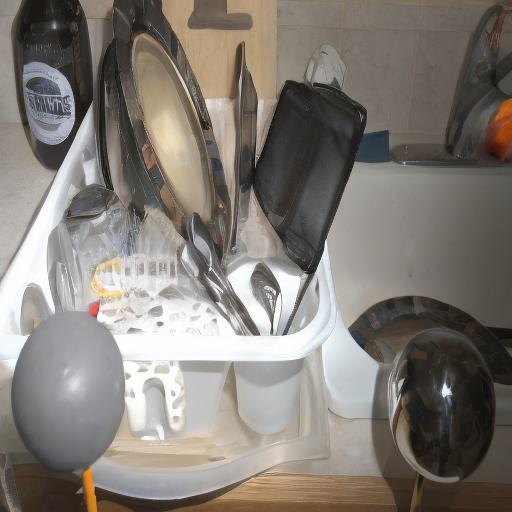} &
\includegraphics[width=0.16\linewidth]{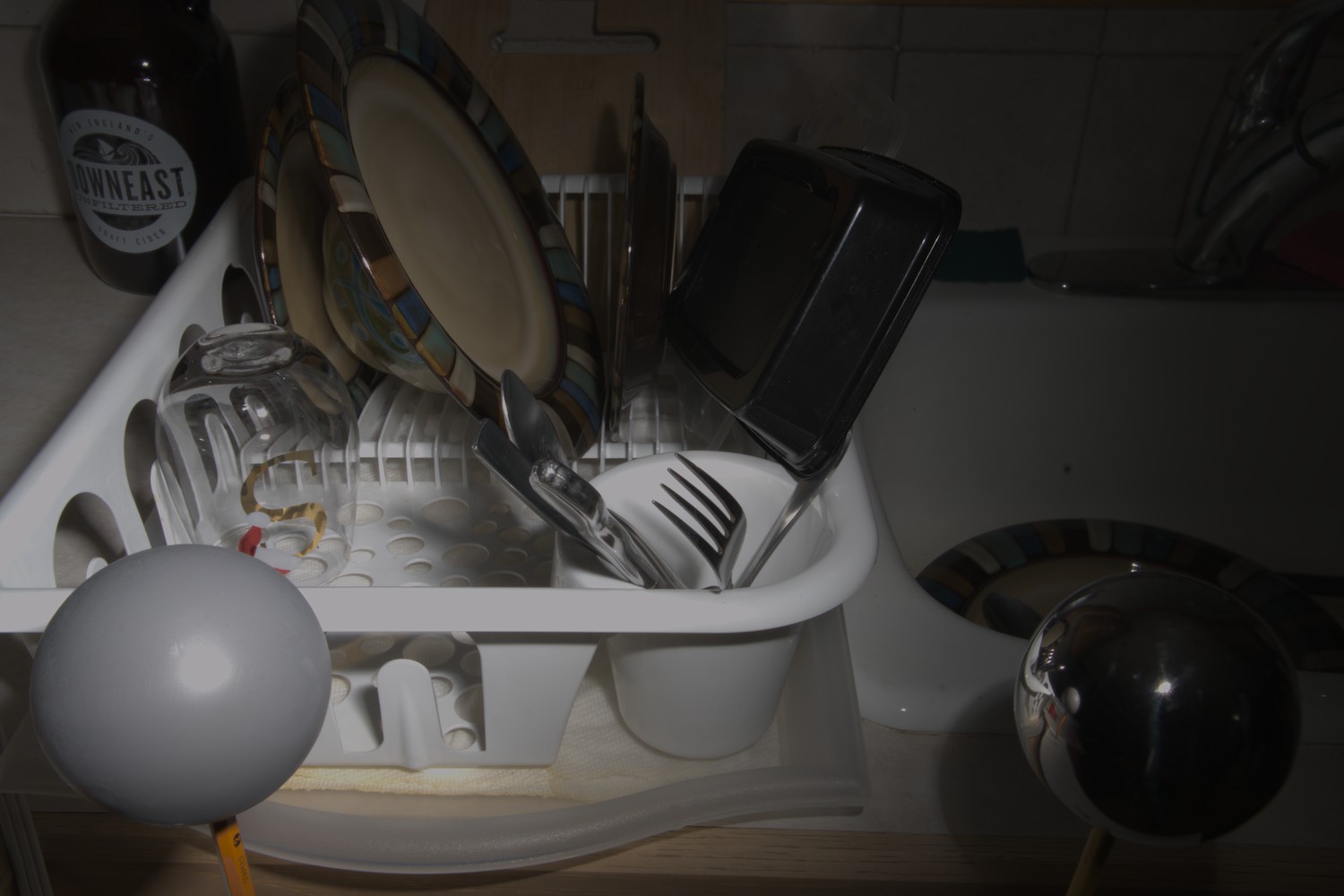} \\

\includegraphics[width=0.16\linewidth]{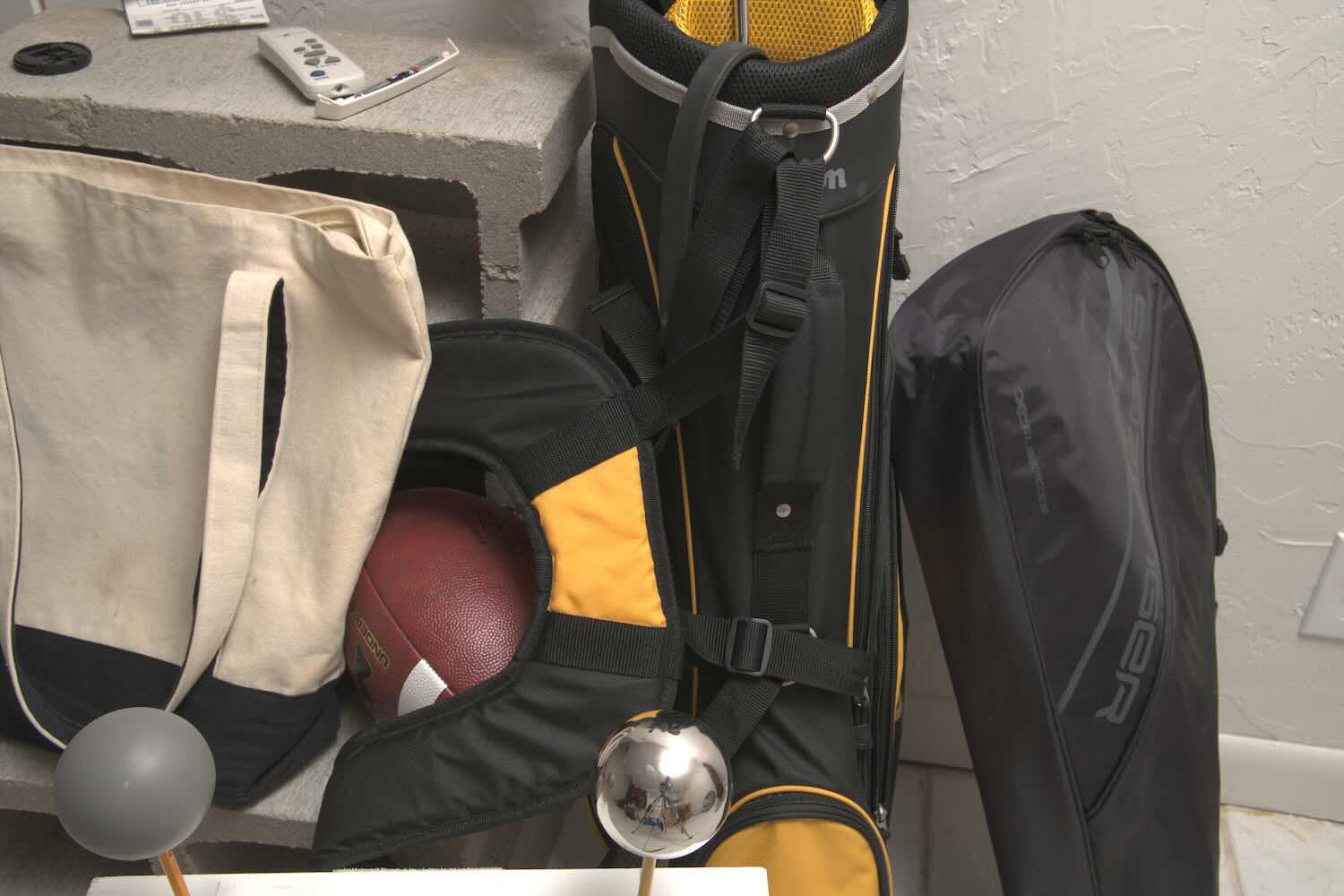} &
\includegraphics[width=0.16\linewidth]{figures/img_src/RADIO_relit/everett_kitchen4/input.jpg} &
\includegraphics[width=0.16\linewidth]{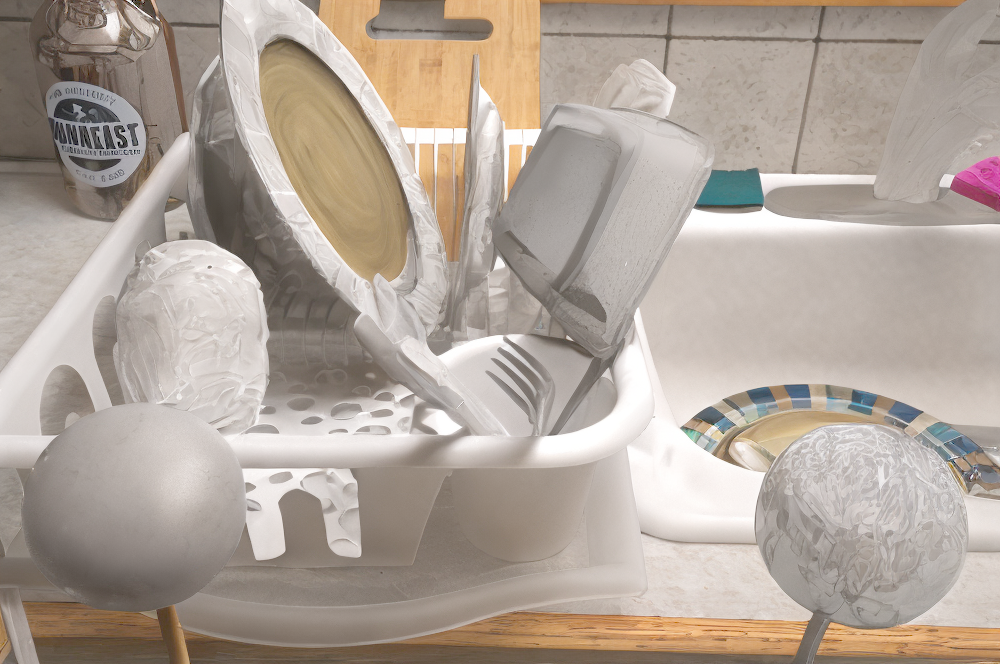} &
\includegraphics[width=0.16\linewidth,height=11.53ex]{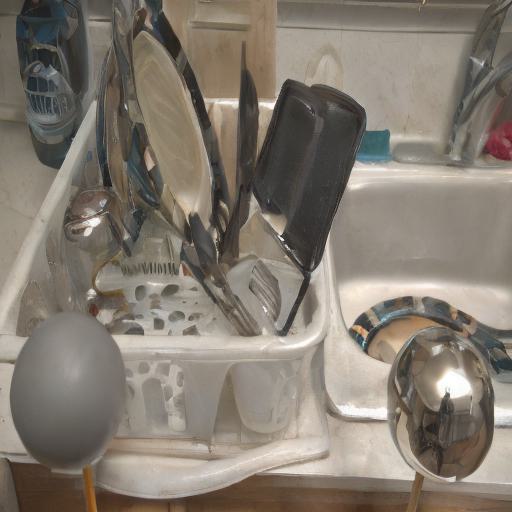} &
\includegraphics[width=0.16\linewidth,height=11.53ex]{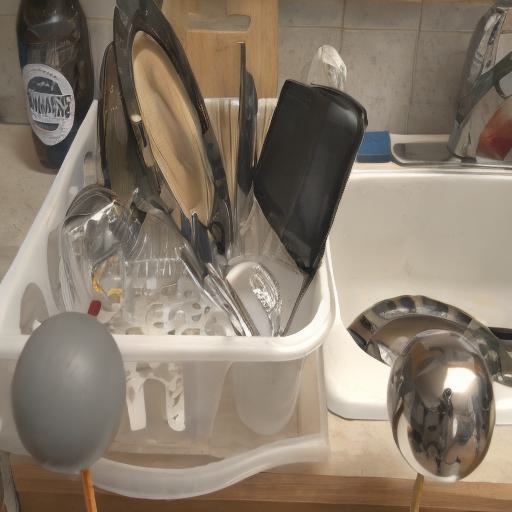} &
\includegraphics[width=0.16\linewidth]{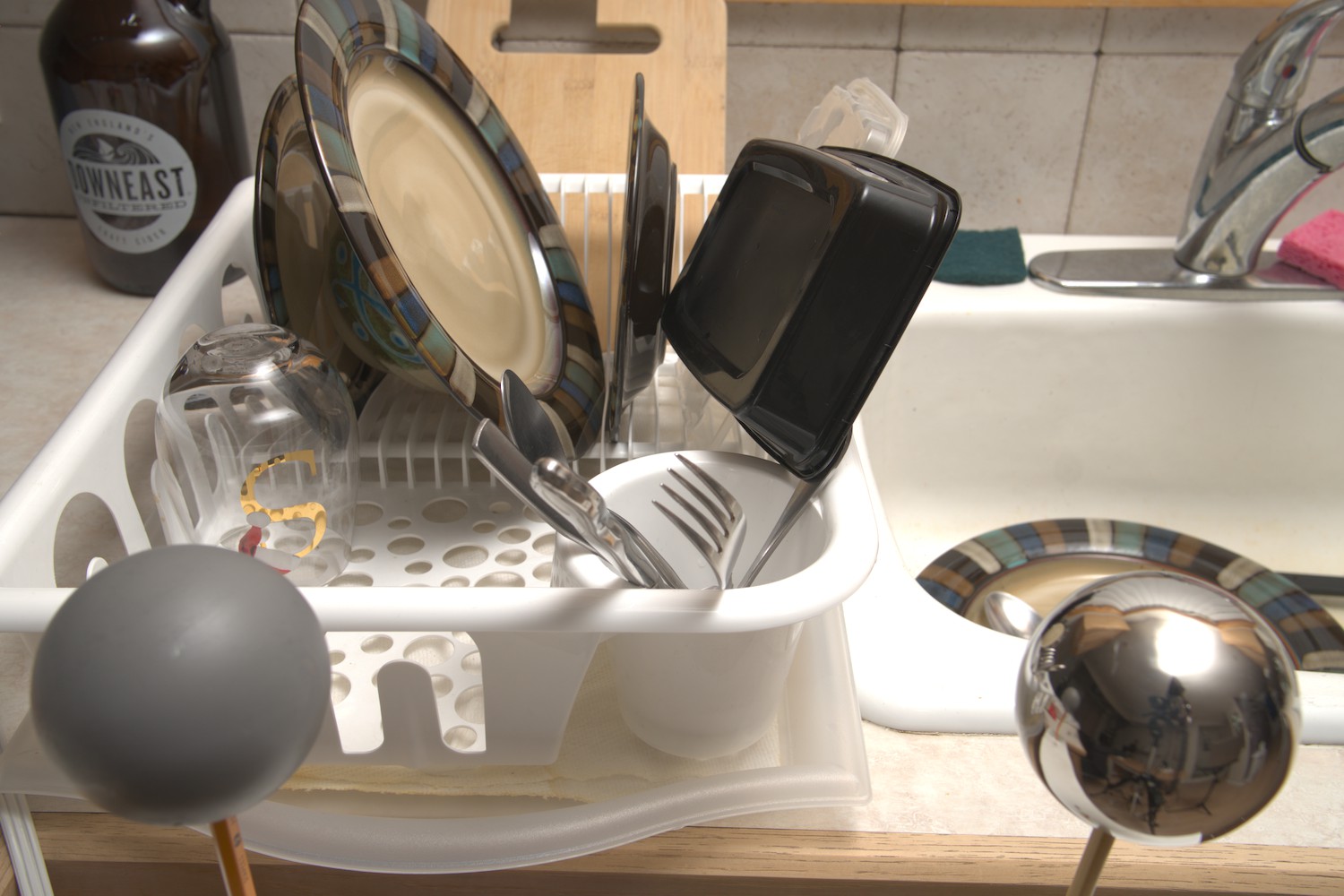} \\

\includegraphics[width=0.16\linewidth]{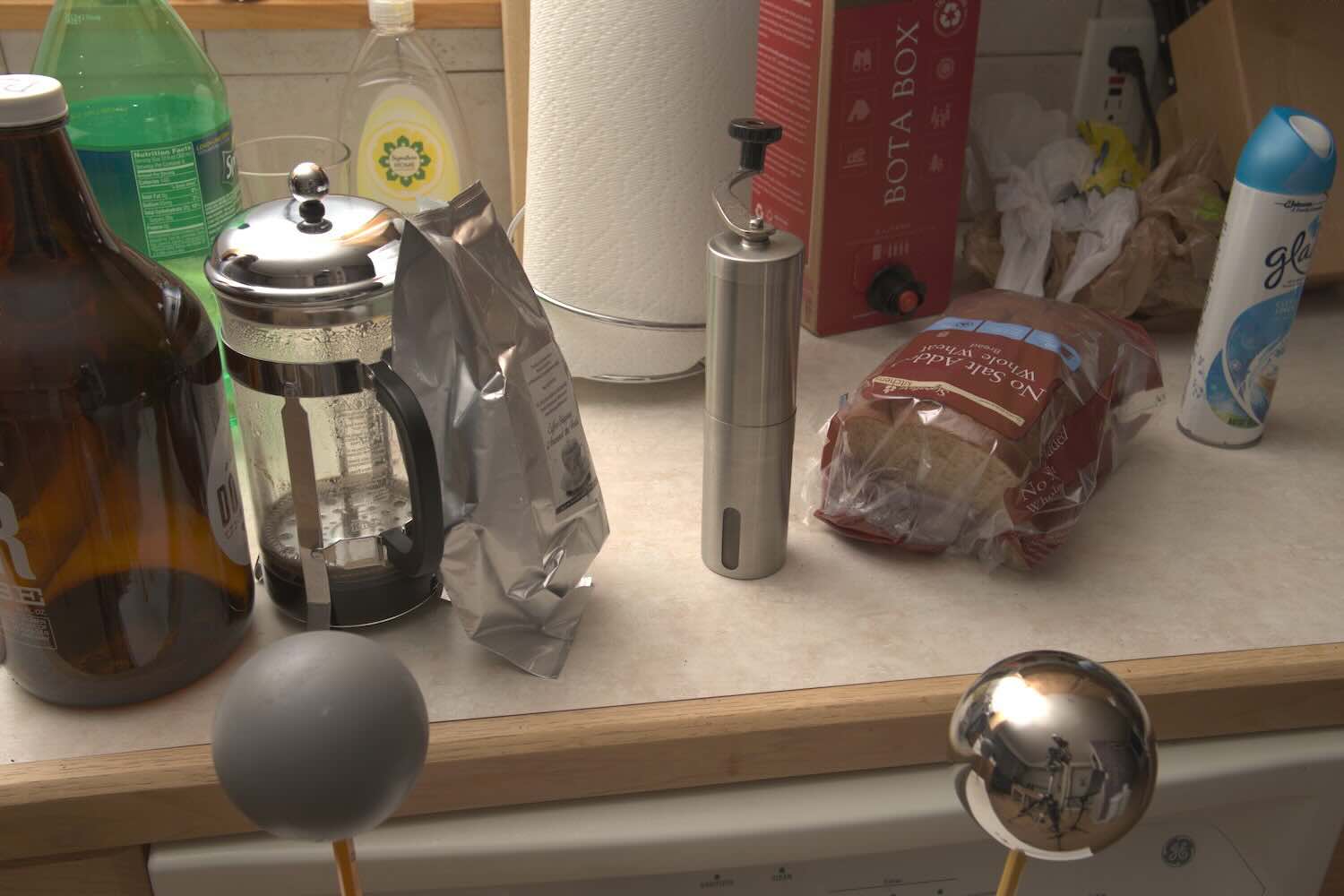} &
\includegraphics[width=0.16\linewidth]{figures/img_src/RADIO_relit/everett_kitchen4/input.jpg} &
\includegraphics[width=0.16\linewidth]{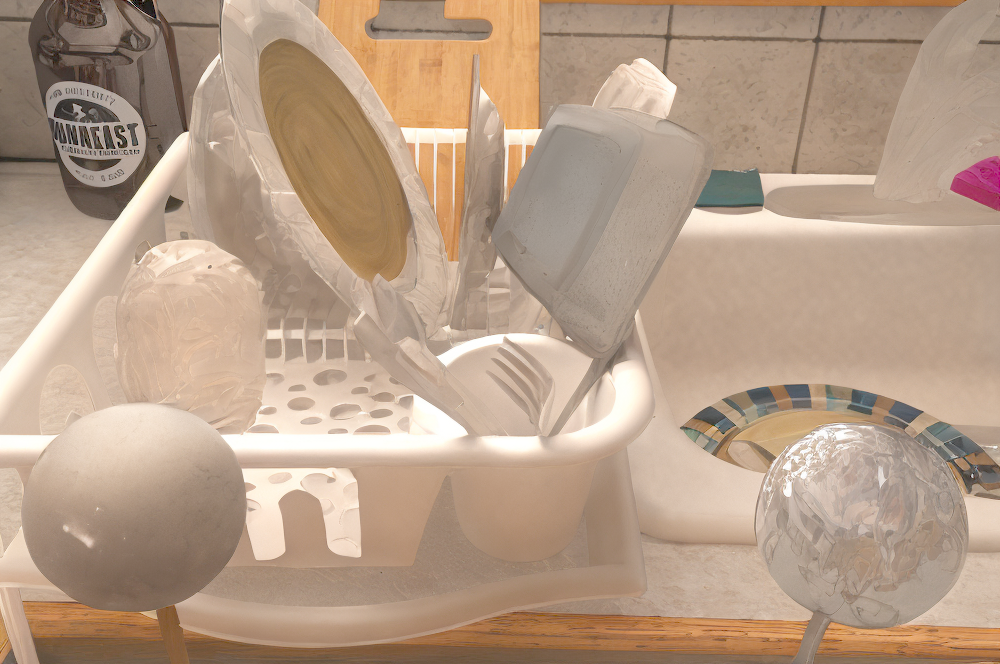} &
\includegraphics[width=0.16\linewidth,height=11.53ex]{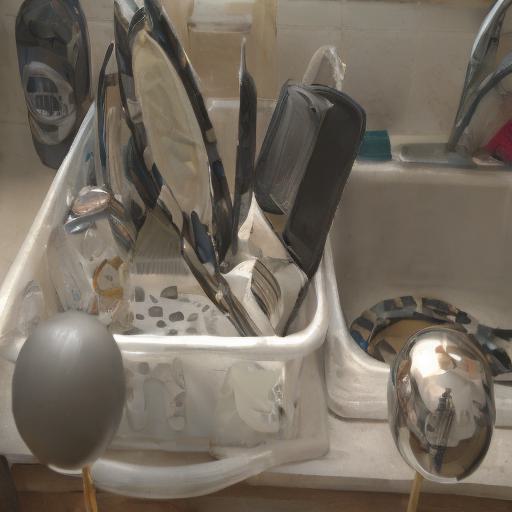} &
\includegraphics[width=0.16\linewidth,height=11.53ex]{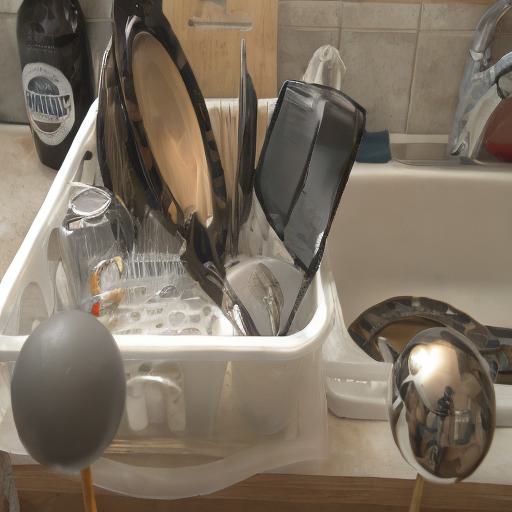} &
\includegraphics[width=0.16\linewidth]{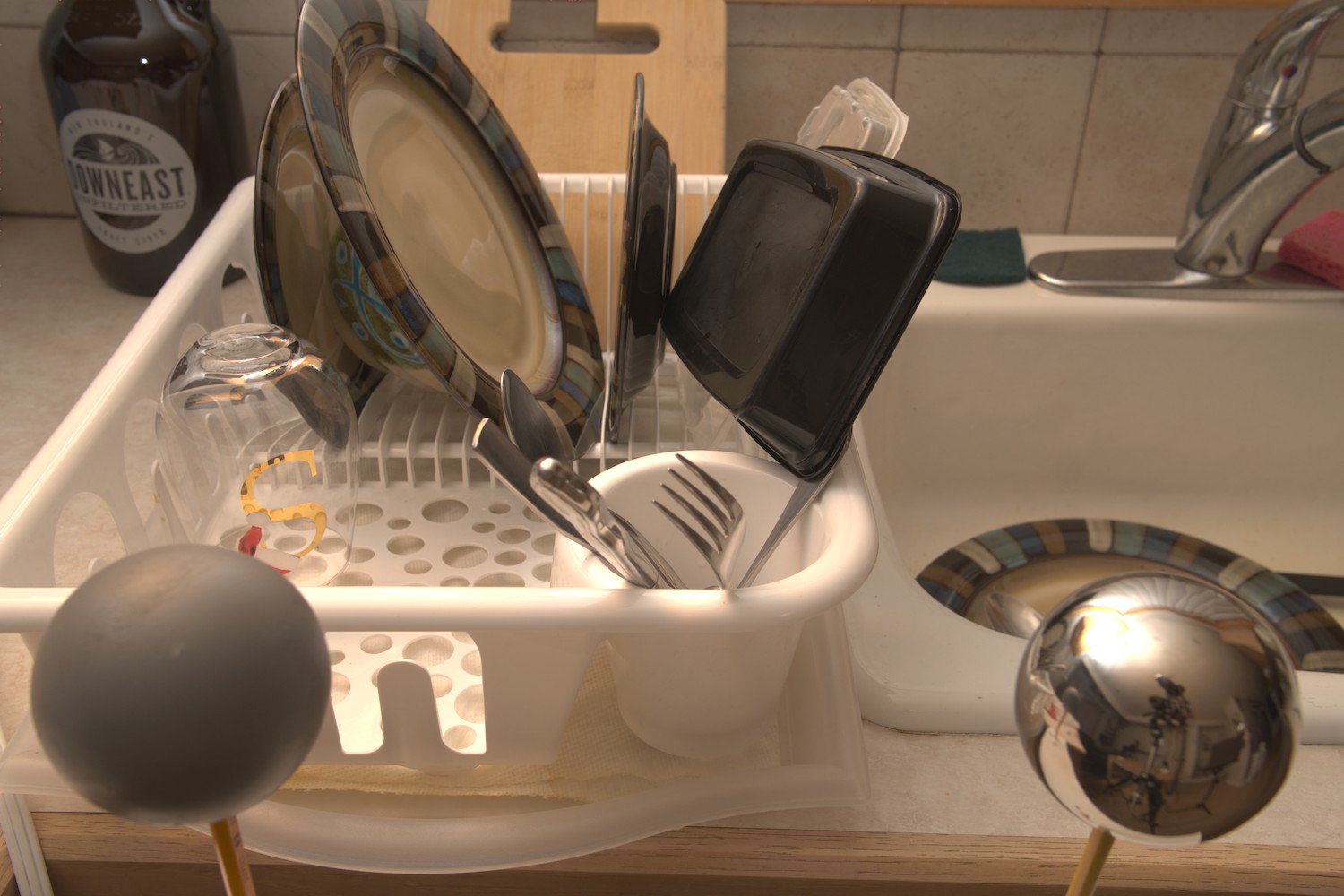} \\

\includegraphics[width=0.16\linewidth]{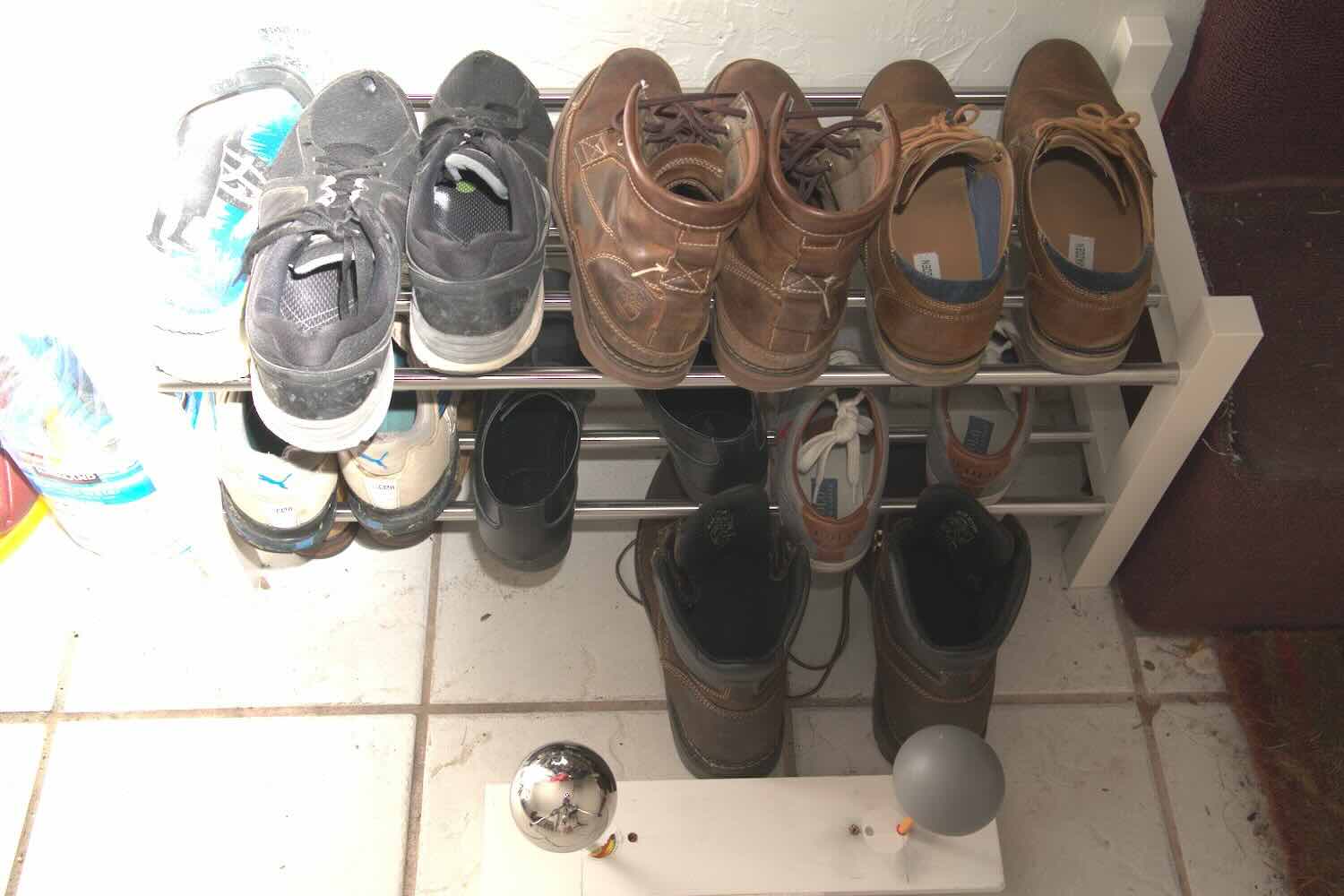} &
\includegraphics[width=0.16\linewidth]{figures/img_src/RADIO_relit/everett_kitchen4/input.jpg} &
\includegraphics[width=0.16\linewidth]{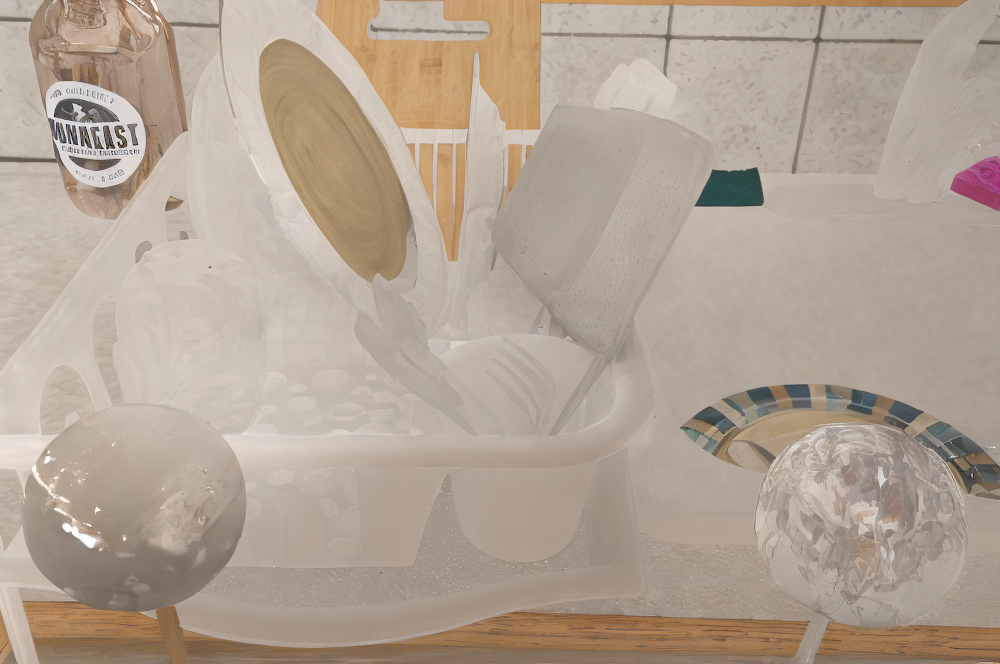} &
\includegraphics[width=0.16\linewidth,height=11.567ex]{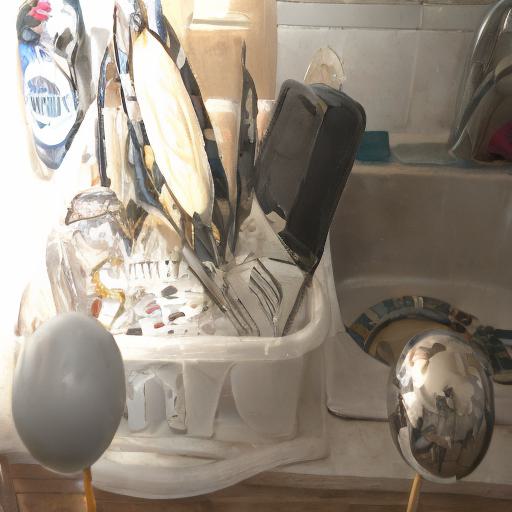} &
\includegraphics[width=0.16\linewidth,height=11.567ex]{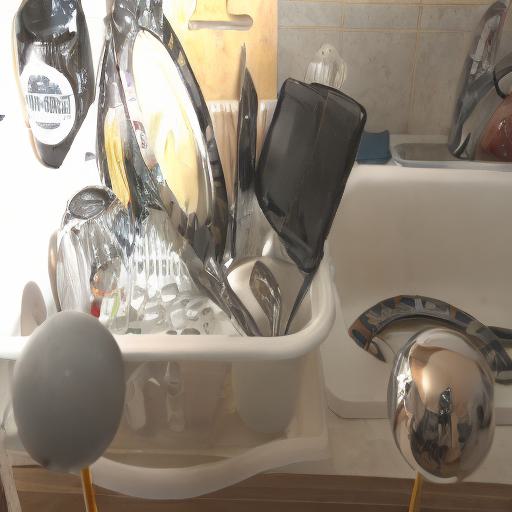} &
\includegraphics[width=0.16\linewidth]{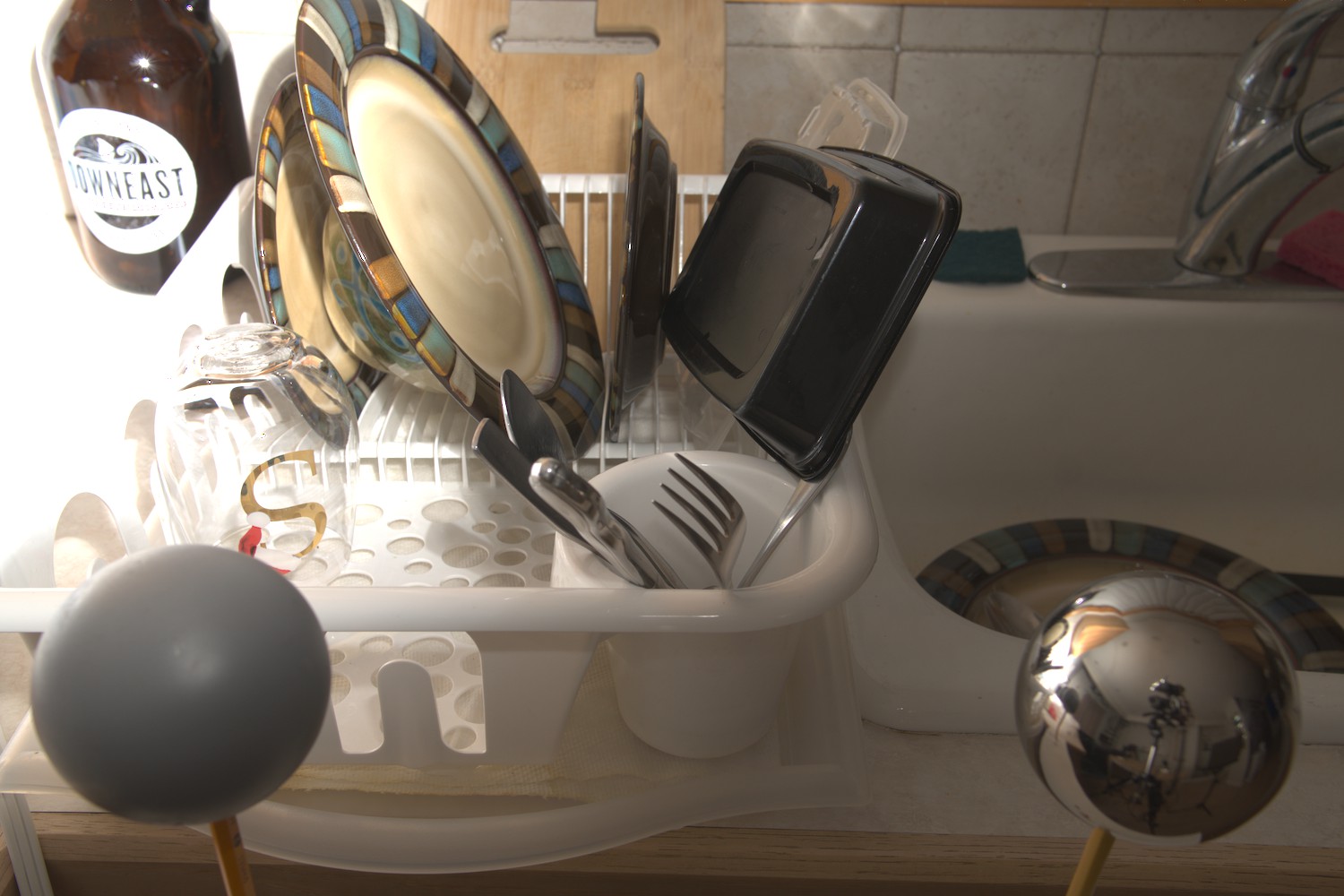}\\
\small Target Light & \small Input & \small RGB-X & \small LumiNet & \small Ours & \small Ground Truth
\end{tabular}
}
\caption{\textbf{Scene relighting comparison on the MIIW~\cite{MIIW} dataset.} Each row corresponds to a different target lighting condition (left). We compare the relighting outputs of RGB-X, LumiNet, and our method against the ground truth {(with bypass decoder disabled)}.}
\label{fig:mit_m_relight}
\end{figure*}

\begin{figure}
    \centering
    \includegraphics[width=\linewidth]{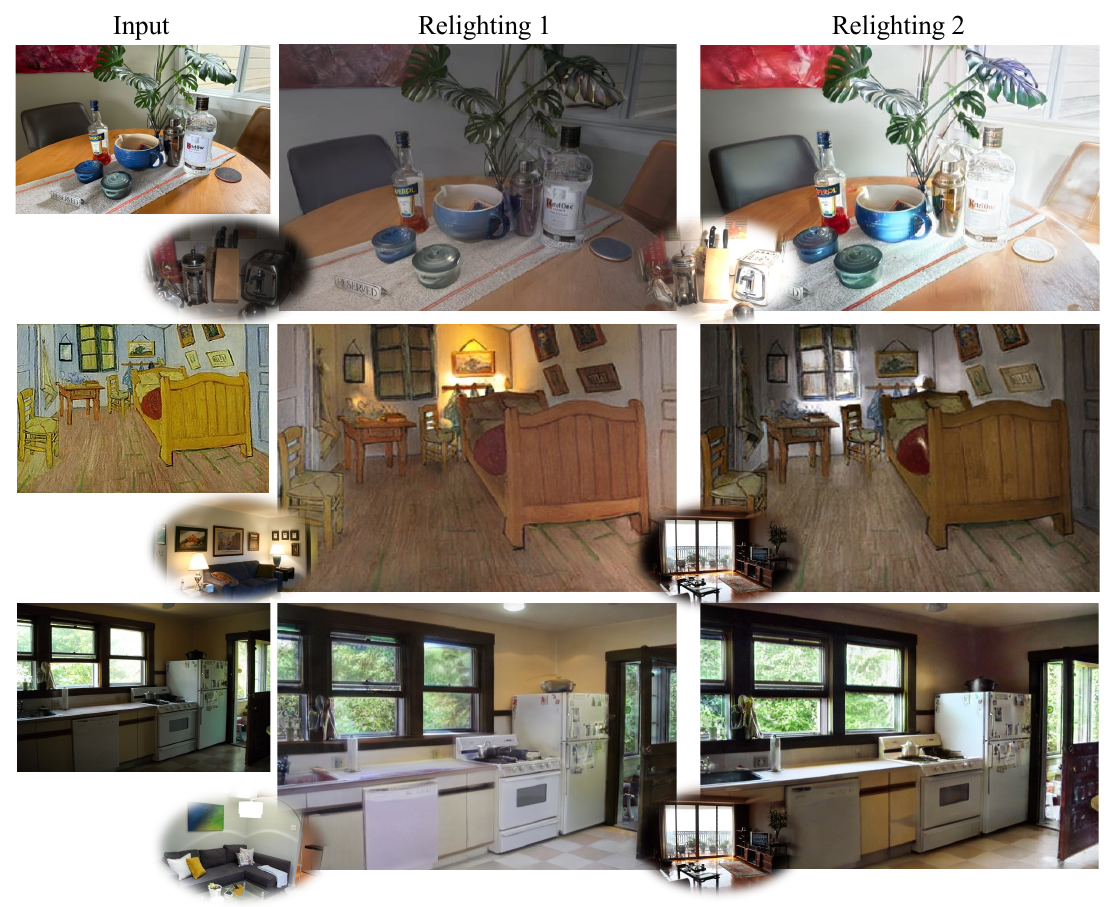}
    \caption{\textbf{In-the-wild relighting.} Our method is capable of relighting a wide variety of scenes while preserving geometry and scene identity. These include: 1) a real indoor image captured by a phone, 2) a painting, and 3) an internet-sourced image. The target lightings are indicated in the bottom left corner of each image.}
    \label{fig:real-world_supp}
\end{figure}

We present additional visual results on the MIIW dataset and in-the-wild images. Fig.~\ref{fig:mit_m_relight} illustrates the same scene relit under different target lighting conditions. Fig.~\ref{fig:real-world_supp} demonstrates relighting on in-the-wild examples. 

Benefiting from the proposed augmented latent-intrinsics feature, our method is capable of relighting a wide variety of scenes while preserving geometry and scene identity. These include: 1) a real indoor image captured by a phone, 2) a painting, and 3) an internet-sourced image.

\begin{table}[]
    \centering
        \caption{\textbf{Mapping of material classes to reflectance clusters.} Each material class is assigned a cluster label based on its dominant reflectance properties: Diffuse, Glossy, Specular, Metallic, or Uncertain. The cluster name is shown only at the start of each group for clarity.}
\begin{tabular}{lll}
\toprule
\textbf{Cluster} & \textbf{Class Index} & \textbf{Class Label} \\
\midrule
Diffuse  & 3  & Cardboard \\
         & 5  & Concrete \\
         & 6  & Cork/corkboard \\
         & 7  & Dirt \\
         & 8  & Fabric/cloth \\
         & 9  & Foliage \\
         & 10 & Food \\
         & 11 & Fur \\
         & 14 & Laminate \\
         & 16 & Linoleum \\
         & 21 & Paper/tissue \\
         & 25 & Sponge \\
         & 26 & Styrofoam \\
         & 29 & Wallpaper \\
         & 31 & Wicker \\
         & 32 & Wood \\
         & 33 & Stone \\
         & 34 & Chalkboard/blackboard \\
         & 35 & Carpet/rug \\
         & 36 & Brick \\
\midrule        
Glossy   & 4  & Ceramic \\
         & 15 & Leather \\
         & 23 & Plastic — opaque \\
         & 24 & Rubber/latex \\
         & 27 & Tile \\
         & 30 & Wax \\
\midrule         
Specular & 12 & Glass \\
         & 18 & Mirror \\
         & 22 & Plastic - clear \\
\midrule
Metallic & 17 & Metal \\
\midrule
Uncertain / Mixed & 0  & unassigned \\
                & 1  & I can't tell \\
                & 2  & More than one material \\
                & 13 & Granite/marble \\
                & 19 & Not on list \\
                & 20 & Painted \\
                & 27 & Tile \\
                & 28 & splitshape \\
                & 37 & Skin \\
                & 38 & Water \\
                & 39 & Hair \\
                & 40 & no\_consensus \\
\bottomrule
\end{tabular}
    \label{tab:mat_class}
\end{table}


\end{document}